\documentclass{llncs}
\usepackage{makeidx}  
\usepackage{times}
\usepackage{helvet}
\usepackage{courier}
\usepackage{tikz}
\usepackage{amssymb}
\usepackage{xspace}
\usepackage{amsmath}
\usepackage{algorithm}
\usepackage{algpseudocode}
\usepackage{adjustbox}

\usepackage{listings}

\newcommand{\todo}[1]{\framebox{{\large\bf TODO:~}{#1}}}
\newcommand{\D}{\ensuremath{D}}

\newcommand{\bigO}{\ensuremath{\mathcal O}}
\newcommand{\matchreg}{\ensuremath{match\mbox -regular}\xspace}

\newcommand{\minsize}{\ensuremath{min\mbox -size}\xspace}
\newcommand{\maxgap}{\ensuremath{max\mbox -gap}\xspace}
\newcommand{\maxspan}{\ensuremath{max\mbox -span}\xspace}

\newcommand{\Maxspan}{\ensuremath{Max\mbox -span}\xspace}
\newcommand{\discriminant}{\ensuremath{discriminant}\xspace}
\newcommand{\posmatch}{\ensuremath{position\mbox -match}\xspace}

\newcommand{\isemb}{\ensuremath{is\mbox -embedding}\xspace}
\newcommand{\Isemb}{\ensuremath{Is\mbox -embedding}\xspace}

\newcommand{\optional}[1]{} 
\newcommand{\subsubspace}{\vspace*{-0.8em}} 
\newcommand{\parspace}{\vspace*{-0.4em}} 

\newcommand{\var}{\bf}
\newcommand{\surl}[1]{{\footnotesize\url{#1}}}
\newcommand{\seq}[1]{{\ensuremath{\mathrm {\langle \lit{#1} \rangle}}}\xspace} 
\newcommand{\lit}[1]{{\ensuremath{\mathrm {\lowercase{#1}}}}\xspace}
\renewcommand{\equiv}{\Longleftrightarrow}
\lstdefinelanguage{zinc} 
  {morekeywords={declarative programming, sequence mining, sequence mining, episode mining, serial episode mining, pattern mining, constraint programming, constrained pattern mining}
  ,classoffset=1
  ,sensitive=false
  ,comment=[l]{\%}
  ,morecomment=[l]{//}
  ,morecomment=[s]{/*}{*/}
  ,morestring=[b]"
  ,literate=
  }

\begin{document}
\title{Constraint-based sequence mining\\using constraint programming\protect\footnote{This paper is published at CPAIOR 2015, this arxiv version additionally has an appendix.}}

\author{Benjamin Negrevergne \and Tias Guns}
\authorrunning{B. Negrevergne and T. Guns} 
\institute{DTAI Research group,\\ KU Leuven\\
  \email{\{firstname\}.\{lastname\}@cs.kuleuven.be}
}

\maketitle
\begin{abstract}

The goal of constraint-based sequence mining is to find sequences of symbols that are included in 
a large number of input sequences and that satisfy some constraints specified by the user. Many constraints have been proposed in the literature, but a general framework is still missing. We investigate the use of constraint programming as general framework for this task.

We first identify four categories of constraints that are applicable to sequence mining. We then propose two constraint programming formulations.  The first formulation introduces a new global constraint called {\em exists-embedding}. This formulation is the most efficient but does not support one type of constraint. To support such constraints, we develop a second formulation that is more general but incurs more overhead. Both formulations can use the projected database technique used in specialised algorithms.


Experiments demonstrate the flexibility towards constraint-based settings and compare the approach to existing methods.

\end{abstract}


\section{Introduction} \label{sec:intro}
In AI in general and in data mining in particular, there is an increasing interest in developing general methods for data analysis. In order to be useful, such methods should be easy to extend with domain-specific knowledge. 

In pattern mining, the frequent sequence mining problem has already been studied in depth, but usually with a focus on efficiency and less on generality and extensibility. An important step in the development of more general approaches was the cSpade algorithm~\cite{zaki2000sequence} which supports a variety constraints.
It supports many constraints such as constraints on the length of the pattern, on the maximum gap in embeddings or on the discriminative power of the patterns between datasets.
Many other constraints have been integrated into specific mining algorithms (e.g. \cite{han2001prefixspan,yan2003clospan,wang2004bide,OhtaniKUA09}).
However, none of these are truly generic in that adding extra constraints usually amounts to changing the data-structures used in the core of the algorithm. 



For \textit{itemset} mining, the simplest form of pattern mining, it has been shown that constraint programming (CP) can be used as a generic framework for constraint-based mining~\cite{cp4im_aij} and beyond~\cite{DBLP:conf/cpaior/RojasBLCL14,dominance_dp}. 
Recent works have also investigated the usage of CP-based approaches for mining sequences with explicit wildcards~\cite{coquery2012sat,JSS-13-3,kemmarictai}. A wildcard represents the presence of exactly one arbitrary symbol in that position in the sequence. 

The main difference between mining itemsets, sequences with wildcards and standard sequences lies in the complexity of testing whether a pattern is included in another itemset/sequence, e.g. from the database. For itemsets, this is simply testing the subset inclusion relation which is easy to encode in CP. For sequences with wildcards and general sequences, one has to check whether an \textit{embedding} exists (matching of the individual symbols). But in case only few embeddings are possible, as in sequences with explicit wildcards, this can be done with a disjunctive constraint over all possible embeddings \cite{kemmarictai}. In general sequence (the setting we address in this paper), a pattern of size $m$ can be embedded into a sequence of size $n$ in $O(n^m)$ different ways, hence prohibiting a direct encoding or enumeration. 


The contributions of this paper are as follows: 
\begin{itemize} 
\item 
      We present four categories of user-constraints, this categorization will be useful to compare the generality of the two proposed models. 
\item We introduce an {\em exists-embedding} global constraint for sequences, and show the relation to projected databases and \textit{projected frequency} used in the sequence mining literature to speedup the mining process \cite{han2001prefixspan,zaki2001spade}. 
\item We propose a more general formulation using a decomposition of the {\em exists-embedding} constraint. Searching whether an embedding exists for each transaction is not easily expressed in CP and requires a modified search procedure. 
\item We investigating the effect of adding constraints, and compare our method with state-of-the-art sequence mining algorithms.
\end{itemize}


\noindent The rest of the paper is organized as follows: Section~\ref{sec:preliminaries} formally introduces the sequence mining problem and the constraint categories. Section~\ref{sec:seq_cp} explains the basics of encoding sequence mining in CP. Section~\ref{sec:first-model} and \ref{sec:second-model} present the model with the global constraint and the decomposition respectively. Section~\ref{sec:experiments} presents the experiments. After an overview of related work (Section~\ref{sec:related}), we discuss the proposed approach and results in Section~\ref{sec:conclusions}.

\section{Sequence mining}
\label{sec:preliminaries}

Sequence mining~\cite{agrawal1995mining} can be seen as a variation of the well-known itemset mining problem proposed in \cite{agrawal1994fast}. In itemset mining, one is given a set of \textit{transactions}, where each transaction is a set of items, and  the goal is to find  patterns (i.e. sets of items) that are included in a large number of transactions.  In sequence mining, the problem is similar except that both transactions and patterns are ordered, (i.e. they are sequences instead of sets) and symbols can be repeated. 
%
For example, $\seq{b,a,c,b}$ and $\seq{a,c,c,b,b}$ are two sequences, and the sequence $\seq{a,b}$ is one possible pattern included in both.

This problem is known in the literature under multiple names, such as {\em embedded subsequence mining}, {\em sequential pattern mining}, {\em flexible motif mining}, or {\em serial episode mining} depending on the application.

\subsection{Frequent sequence mining: problem statement}
\label{sec:freq-sequ-mining}
A key concept of any pattern mining setting is the pattern inclusion relation. 
In sequence mining, a pattern is included in a transaction if there exists an embedding of that sequence in the transaction; where an embedding is a mapping of every symbol in the pattern to the same symbol in the transaction such that the order is respected. 

\begin{definition}[Embedding in a sequence]
  \label{def:seq-embedding}
  Let $S = \langle s_1,\ldots, s_m \rangle$ and $S' = \langle s'_1, \ldots, s'_n\rangle$ be two sequences of size $m$ and $n$ respectively with $m\leq n$. The tuple of integers $e = (e_1, \ldots, e_m)$ is an \textbf{embedding} of $S$ in $S'$  (denoted $S \sqsubseteq_e S'$) if and only if:
  \begin{align}
    S \sqsubseteq_e S' \leftrightarrow e_1 < \ldots < e_m ~\mbox{and}~ \forall i \in 1,\ldots,m: s_i = s'_{e_i}
  \end{align}
\end{definition}
For example, let $S=\seq{a,b}$ be a pattern, then $(2,4)$ is an embedding of $S$ in $\seq{b,a,c,b}$ and $(1,4),(1,5)$ are both embeddings of $S$ in $\seq{a,c,c,b,b}$. An alternative 
setting considers sequences of \textit{itemsets} instead of sequences of individual symbols. In this case, the definition is $S \sqsubseteq_e S' \leftrightarrow e_1 < \ldots < e_n ~\mbox{and}~ \forall i \in 1,\ldots,n: s_i \subseteq s'_{e_i}$. We do not consider this setting further in this paper, though it is an obvious extension.

We can now define the sequence inclusion relation as follows:
\begin{definition}[Inclusion relation for sequences]
  \label{def:seq-incl-rel}
  Given two sequences $S$ and $S'$, $S$ \textbf{is included in} $S'$ (denoted $S \sqsubseteq S'$) if there exists an embedding $e$ of $S$ in $S'$:
  \begin{align} \label{eq:sec-incl-rel}
    S \sqsubseteq S' \leftrightarrow \exists e ~\mbox{s.t.}~ S \sqsubseteq_e S'.
  \end{align}
\end{definition}
To continue on the example above, $S=\seq{a,b}$ is included in both $\seq{b,a,c,b}$ and $\seq{a,c,c,b,b}$ but not in $\seq{c,b,a,a}$.

\begin{definition}[Sequential dataset]
  Given an alphabet of symbols $\Sigma$, a {\em sequential dataset} $D$ is a multiset of sequences defined over symbols in $\Sigma$. 
\end{definition}
Each sequence in $D$ is called a {\em transaction} using the terminology from itemset mining. The number of transactions in $D$ is denoted $|D|$ and the sum of the lengths of every transaction in $D$ is denoted $||D||$ ($||D|| = \sum_{i = 1}^{|D|}|T_i|$).
Furthermore, we use {\em dataset} as a shorthand for {\em sequential dataset} when it is clear from context.

Given a dataset $D = \{T_i, \ldots, T_n\}$, one can compute the {\bf cover} of a sequence $S$ as the set of all transactions $T_i$ that contain $S$:
\begin{equation}
  \label{eq:cover}
cover(S, \D) = \{T_i \in \D : S \sqsubseteq T_i\}
\end{equation}

We can now define frequent sequence mining, where the goal is to find all patterns that are frequent in the database; namely, the size of their cover is sufficiently large.

\begin{definition}[Frequent sequence mining]
\label{def:spm}
Given:
\begin{enumerate}
\item an alphabet $\Sigma$ 
\item a sequential dataset $\D = \{T_1, \ldots, T_{n}\}$ defined over $\Sigma$ 
\item a minimum frequency threshold $\theta$,
\end{enumerate}
\noindent enumerate all sequences $S$ 
such that
$
  |cover(S, \D)| \ge \theta  
$.
\end{definition}%

In large datasets, the number of frequent sequences is often too large to be analyzed by a human. Extra constraints can be added to extract fewer, but more relevant or interesting patterns. Many such constraints have been studied in the past.

\subsection{Constraints}
\label{sec:constraints-max-1}

Constraints typically capture background knowledge and are provided by the user. We identify four categories of constraints for sequence mining: 1) constraints over the pattern, 2) constraints over the cover set, 3) constraints over the inclusion relation and 4) preferences over the solution set. 

\subsubspace
\subsubsection{Constraints on the pattern}
These put restrictions on the structure of the pattern. Typical examples include size constraints 
or regular expression constraints.\\
{\em Size constraints:}
A size constraint is simply $|S| \gtrless \alpha$ where $\gtrless \in \{=,\neq,>,\geq,<,\leq\}$ and $\alpha$ is a user-supplied threshold. It is used to discard  small patterns.\\
%
{\em Item constraints:} One can constrain a symbol $t$ to surely be in the pattern: $\exists s \in S: s = t$; or that it can not appear in the pattern: $\forall s \in S: s \neq t$, or more complex logical expressions over the symbols in the pattern.\\
{\em Regular expression constraints:}
Let $R$ be a regular expression over the vocabulary $V$ and $L_R$ be the language of sequences recognised by $R$, then for any sequence pattern $S$ over $V$, the {\em \matchreg} constraint requires that $S \in L_R$~\cite{han2001prefixspan}.

\subsubspace
\subsubsection{Constraints on the cover set.}
The {\em minimum frequency} constraint $|cover(S,D)| \geq \theta$ is the most common example of a constraint over the cover set. Alternatively, one can impose the {\em maximum frequency} constraint: $|cover(S,D)| \leq \beta$\\ 
{\em Discriminating constraints:} In case of multiple datasets, discriminating constraints require that patterns effectively distinguish the datasets from each other. 
Given two datasets $D_1$ and $D_2$, one can require that the ratio between the size of the cover of both is above a threshold: $\frac{|cover(S,D_1)|}{|cover(S,D_2)|} \geq \alpha$. Other examples include more statistical measures such as information gain and entropy~\cite{correlated_cp}.


\subsubspace
\subsubsection{Constraints over the inclusion relation.}
The inclusion relation in definition~\ref{def:seq-incl-rel} states that $S \sqsubseteq S' \leftrightarrow \exists e ~\mbox{s.t.}~ S \sqsubseteq_e S'$. Hence, an embedding of a pattern can match symbols that are far apart in the transaction. For example, the sequence $\seq{a,c}$ is embedded in the transaction $\seq{a,b,b,b,\ldots,b,c}$ independently of the distance between $\lit a$ and $\lit c$ in the transaction. This is undesirable when mining datasets with long transactions. The \maxgap and \maxspan constraints  \cite{zaki2000sequence}  impose a restriction on the embedding, and hence on the inclusion relation. 
\label{sec:maxgap-constraint}
The {\em \maxgap constraint} is satisfied on a transaction $T_i$ if an embedding $e$ maps every two consecutive symbols in $S$ to symbols in $T_i$ that are close to each-other: $\maxgap_i(e) \Leftrightarrow \forall{j \in 2..|T_i|}, (e_{j} - e_{j-1} - 1) \le \gamma$.
For example,  the sequence $\seq{abc}$ is embedded in the transaction \seq{adddbc} with a maximum gap of 3 whereas \seq{ac} is not.
%
\label{sec:maxspan-constraint}
The {\em \maxspan constraint} requires that the distance between the first and last position of the embedding of all transactions $T_i$ is below a 
threshold~$\gamma$: 
$
  \maxspan_i(e) \Leftrightarrow e_{|T_i|} - e_1 + 1 \le \gamma
$.

\subsubspace
\subsubsection{Preferences over the solution set.}
A pairwise preference over the solution set expresses that a pattern $A$ is preferred over a pattern $B$. In~\cite{dominance_dp} it was shown that condensed representations like closed, maximal and free patterns can be expressed as pairwise preference relations. Skypatterns~\cite{DBLP:conf/cpaior/RojasBLCL14} and multi-objective optimisation can also be seen as preference over patterns. As an example, let $\Delta$ be the set of all patterns; then, the set of all closed patterns is $\{S \in \Delta | \nexists S' \mbox{ s.t. } S \sqsubset S' \mbox{ and } cover(S,\D) = cover(S',\D)\}$.






\section{Sequence Mining in Constraint Programming}\label{sec:seq_cp}
In constraint programming, problems are expressed as a constraint satisfaction problem (CSP), or a constraint optimisation problem (COP). A CSP $X=(V,D,C)$ consists of a set of variables $V$, a finite domain $D$ that defines for each variable $v \in V$ the possible values that it can take, and a set of constraints $C$ over the variables in $V$.  A solution to a CSP is an assignment of each variable to a value from its domain such that all constraints are satisfied. A COP additionally consists of an optimisation criterion $f(V)$ that expresses the quality of the solution. 

There is no restriction on what a constraint $C$ can represent.  Examples include logical constraints like ${\var X} \wedge {\var Y}$ or ${\var X} \rightarrow {\var Y}$ and mathematical constraints such as ${\var Z} = {\var X} + {\var Y}$ etc. Each constraint has a corresponding \textit{propagator} that ensures the constraint is satisfied during the search. Many \textit{global constraints} have been proposed, such as \textit{alldifferent}, which have a custom propagator that is often more efficient then if one would \textit{decompose} that constraint in terms of simple logical or mathematical constraints. A final important concept used in this paper is that of \textit{reified constraints}. A reified constraint is of the form ${\var B} \leftrightarrow C'$ where ${\var B}$ is a Boolean variable which will be assigned to the truth value of constraint $C'$. Reified constraints have their own propagator too.

\parspace
\paragraph{Variables and domains for modeling sequence mining.}
Modeling a problem as a CSP requires the definition of a set of variables with a finite domain, and a set of constraints. One solution to the CSP will correspond to one pattern, that is, one frequent sequence.

We model the problem using an array ${\var S}$ of integer variables representing the characters of the sequence and an array ${\var C}$ of Boolean variables representing which transactions include the pattern. This is illustrated in Fig. \ref{fig:principle1}:
%
\begin{figure}[t]
  \centering
  \Large
      \begin{tikzpicture}[scale=0.55, transform shape]
        \tikzstyle{seq_item}=[draw, fill=blue!20, shape=rectangle, minimum size=1cm];
        \tikzstyle{seq_item_small}=[draw, fill=blue!20, shape=rectangle, minimum size=1cm, minimum width=0.5cm];
        \tikzstyle{seq_item_static}=[draw, fill=white, shape=rectangle, minimum size=1cm];
        \tikzstyle{myedge}=[thick, -latex, color=red!80!black];
        \newcounter{y}

        \node at (0.2, 0) {${\var S}:~$}; 
        \setcounter{y}{1}
        \foreach \i in {A, B, $\epsilon$, $\epsilon$}{
          \node[seq_item, label=above:{\small p=\arabic{y}}] at (\arabic{y}, 0) (j\arabic{y}) {\i};
          \stepcounter{y}
        }

        \node[seq_item_small] (t1) at (-1.1, -1.4) {$1$};
        \node at (-1.8, -1.4) {${\var C_1}:~$};
                
        \node at (0.15, -1.4) {$T_1:~$}; 
        \setcounter{y}{1}
        \foreach \i in {A, C, B}{
          \node[seq_item_static] at (0+\arabic{y}, -1.4) (x1\arabic{y}) {\i};
          \stepcounter{y}
        }
                
        \node[seq_item_small] (t2) at (-1.1, -2.4) {$0$};
        \node at (-1.8, -2.4) {${\var C_2}:~$};
                
        \node at (0.15, -2.4) {$T_2:~$}; 
        \setcounter{y}{1}
        \foreach \i in {B, A, A, C}{
          \node[seq_item_static] at (0+\arabic{y}, -2.4) (x2\arabic{y}) {\i};
          \stepcounter{y}
        }
               
        \end{tikzpicture}
\caption{Example assignment; blue boxes represent variables, white boxes represent data.}
\label{fig:principle1}  
\end{figure}
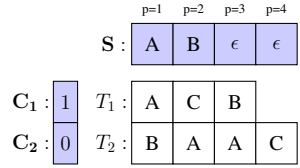%
\begin{enumerate}
\item $T_1$ and $T_2$ represent two transactions given as input. We denote the number of transactions by $n$;
\item The array of variables ${\var S}$ represents the sequence pattern.
  Each variable ${\var S_j}$ represents the character in the $j$th position of the sequence.
  The size of ${\var S}$ is determined by the length of the longest transactions (in the example this is $4$).
  We want to allow patterns that have fewer than $max_i(|T_i)$ characters, hence we use $\epsilon$ to represent an unused position in ${\var S}$.
  The domain of each variable ${\var S_j}$ is thus $\Sigma \cup \{\epsilon\}$;
\item Boolean variables ${\var C_i}$ represent whether the pattern is included in transaction $T_i$, that is, whether ${\var S} \sqsubseteq T_i$. In the example, this is the case for $T_1$ but not for $T_2$.
\end{enumerate}

What remains to be defined is the constraints. The key part here is how to model the inclusion relation; that is, the constraint that verifies whether a pattern is included in the transaction. Conceptually, this is the following reified constraint: ${\var C_i} \leftrightarrow \exists e ~\mbox{s.t.}~ {\var S} \sqsubseteq_e T_i$.
As mentioned in the introduction, the number of possible embeddings is exponential in the size of the pattern. Hence, one can not model this as a disjunctive constraint over all possible embeddings (as is done for sequences with explicit wildcards~\cite{kemmarictai}).

We propose two approaches to cope with this problem: one with a global constraint that verifies the inclusion relation directly on the data, 
and one in which the inclusion relation is decomposed and the embedding is exposed through variables.

\section{Sequence mining with a global \textit{exists-embedding} constraint}
\label{sec:first-model}

The model consists of three parts: encoding of the pattern, of the minimum frequency constraint and finally of the inclusion relation using a global constraint.

\parspace
\paragraph{Variable-length\ pattern:} The array ${\var S}$ has length $k$; patterns with $l < k$ symbols are represented with $l$ symbols from $\Sigma$ and $(k - l)$ times an $\epsilon$ value. To avoid enumerating the same pattern with $\epsilon$ values in different positions, $\epsilon$ values can only appear at the end: 
  \begin{equation}
    \label{eq:well-formed}
    \begin{gathered}
          \forall j \in 1..(k-1): {\var S_{j}} = \epsilon \rightarrow {\var S_{j+1}} = \epsilon
    \end{gathered}
  \end{equation}
\optional{
This can be encoded with $k-1$ auxiliary variables and reified ${\var B} \leftrightarrow {\var S_j} = \epsilon$ constraints, and $1$ lexicographic 'less than or equal' constraint over the array of auxiliary variables (using the observation that ${\var B_1} \rightarrow {\var B_2} \equiv {\var B_1} \leq {\var B_2}$ for Boolean variables).
}

\parspace
\parspace
\paragraph{Minimum frequency:} At least $\theta$ transactions should include the pattern. This inclusion is indicated by the array of Boolean variables ${\var C}$:
\parspace
  \begin{equation}
    \label{eq:minfreq}
    \begin{gathered}
        \sum_{i = 1}^n {\var C_i} \geq \theta
    \end{gathered}
  \end{equation}
\optional{
This can be encoded with $1$ linear inequality constraint over the ${\var C}$ variables.
}

\parspace
\parspace
\paragraph{Global exists-embedding constraint:}
The goal is to encode the relation:
$
{\var C_i} \leftrightarrow \exists e ~\mbox{s.t.}~ {\var S} \sqsubseteq_e T_i.
$
\begin{algorithm}[t]
\algtext*{EndIf}
  \footnotesize 
\caption{Incremental propagator for ${\var C_i} \leftrightarrow \exists e ~\mbox{s.t.}~ {\var S} \sqsubseteq_e T_i$:\label{alg:global_emp}}
\textit{internal state, $pos_S$: current position in ${\var S}$ to check, initially 1}\\
\textit{internal state, $pos_e$: current position in $T_i$ to match to, initially 1}
\begin{algorithmic}[1]
\While{$pos_S \leq |T_i|$ and ${\var S}$[$pos_S$] is assigned\label{a1:while}} \Comment{note that $|T_i| \leq |{\var S}|$} 
\If{${\var S}$[$pos_S$] $\neq \epsilon$}\label{a1:neq_eps}
\While{not ($T_i[pos_e] = {\var S}[pos_S])$ and $pos_e \leq |T_i|$\label{a1:while2}} \Comment find match
\State{$pos_e \gets pos_e + 1$}
\EndWhile
\If{$pos_e \leq |T_i|$} \Comment match found, on to next one
\State{$pos_S \gets pos_S + 1$; $pos_e \gets pos_e + 1$}
\Else
\State propagate ${\var C_i} = False$ and return
\EndIf
\Else \Comment previous ones matched and rest is $\epsilon$
\State propagate ${\var C_i} = True$ and return  \label{a1:eps}
\EndIf
\EndWhile \label{a1:endwhile}
\If{$pos_S > |{\var S}|$} \Comment previous ones matched and reached end of sequence
\State propagate ${\var C_i} = True$ and return  \label{a1:eos}
\EndIf
\If{$pos_S > |T_i|$ and $|T_i| < |{\var S}|$} \label{a1:longer:start}
\State{{\bf let} $R \gets {\var S}[|T_i|+1]$}
\If{$R$ is assigned and $R = \epsilon$} \Comment{{\var S} should not be longer than this transaction}
\State propagate ${\var C_i} = True$ and return 
\EndIf
\If{$\epsilon$ is not in the domain of $R$} 
\State propagate ${\var C_i} = False$ and return
\EndIf
\EndIf \label{a1:longer:stop}
\If{${\var C_i}$ is assigned and ${\var C_i} = True$} \label{a1:revprop:start}
\State{propagate by removing from ${\var S}[pos_S]$ all symbols not in $\langle T_i[pos_e]..T_i[|T_i|] \rangle$ except $\epsilon$}
\EndIf \label{a1:revprop:stop}
\end{algorithmic}
\end{algorithm}%
The propagator algorithm for this constraint is given in Algorithm~\ref{alg:global_emp}. It is an incremental propagator that should be run when one of the ${\var S}$ variables is assigned. Line~\ref{a1:while} will loop over the 
variables in ${\var S}$ until reaching an unassigned one at position $pos_S$. In the sequence mining literature, the sequence $\langle {\var S_{1}}..{\var S_{{pos_S}}} \rangle$ is called the \textit{prefix}. For each assigned ${\var S_j}$ variable, a matching element in the transaction is sought, starting from the position $pos_e$ after the element that matched the previous ${\var S_{j-1}}$ assigned variable. If no such match 
is found then an embedding can not be found and ${\var C_i}$ is set to false.

Line~\ref{a1:eps} is called when an ${\var S_j}$ variable is assigned to $\epsilon$. This line can only be reached if all previous values of ${\var S}$ are assigned and were matched in $T_i$, hence the propagator can set ${\var C_i}$ to true and quit. Similarly for line~\ref{a1:eos} when the end of the sequence is reached, and lines~\ref{a1:longer:start}-\ref{a1:longer:stop} in case the transaction is smaller than the sequence. Lines~\ref{a1:revprop:start}-\ref{a1:revprop:stop} propagate the remaining possible symbols from $T_i$ to the first unassigned ${\var S}$ variable in case ${\var C_i} = True$.

The propagator algorithm has complexity $O(|T_i|)$: the loop on line~\ref{a1:while} is run up to $|T_i|$ times and on line~\ref{a1:while2} at most $|T_i|$ times in total, as $pos_e$ is monotonically increasing.
\optional{
There are $n$ global \textit{exists-embedding} constraints needed.
}

\subsection{Improved pruning with \textit{projected frequency}}
%
Compared to specialised sequence mining algorithms, $pos_S$ 
in Algorithm~\ref{alg:global_emp} points to the first position in ${\var S}$ after the current \textit{prefix}. Dually, $pos_e$ points to the position after the first match of the prefix in the transaction. If one would project the prefix away, only the symbols in the transaction from $pos_e$ on would remain; this is known as \textit{prefix projection}~\cite{han2001prefixspan}. Given prefix $\seq{a,c}$ and transaction $\seq{b,a,a,e,c,b,c,b,b}$ the projected transaction is $\seq{b,c,b,b}$.

The concept of a prefix-projected database can be used to recompute the frequency of all symbols in the projected database. If a symbol is present but not frequent in the projected database, one can avoid searching over it. This is known to speed up specialised mining algorithms considerably~\cite{han2001prefixspan,wang2004bide}.

To achieve this in the above model, we need to adapt the global propagator so that it exports the symbols that still appear after $pos_e$. 
We introduce an auxiliary integer variable ${\var X_i}$ for every transaction $T_i$, whose domain represents these symbols (the set of symbols is monotonically decreasing). To avoid searching over infrequent symbols, we define a custom search routine (brancher) over the ${\var S}$ variables. It first computes the local frequencies of all symbols based on the domains of the ${\var X_i}$ variables; symbols that are locally infrequent will not not be branched over. See Appendix~\ref{app:branch}
for more details.

\subsection{Constraints}
\label{sec:user-constraints1}
This formulation supports a variety of constraints, namely on the pattern (type 1), on the cover set (type 2) and over the solution set (type 4).
For example, the type 1 constraint \minsize, constrains the size of the pattern to be larger than a user-defined threshold~$\alpha$. This constraint can be formalised as follows. 
  \begin{equation}
    \label{eq:minsize}
    \begin{gathered}
      \sum_{j = 1}^k\var \left[S_j \neq \epsilon\right] \ge \alpha
    \end{gathered}
  \end{equation}

\optional{Alternatively, we can use of the $\var B$ auxiliary variables defined in the previous section to simplify this formulation. By observing that $\var B_j \leftrightarrow S_j = \epsilon$,  Equation~\ref{eq:minsize} can be reformulated as: $(\sum_{j = 1}^k\lnot \var B_j) \ge \alpha$.}

{\em Minimum frequency} in Equation~(\ref{eq:minfreq}) is an example of a constraint of type~2, over the cover set. Another example is the {\it discriminative} constraint mentioned in Section~\ref{sec:constraints-max-1}: given two datasets $D_1$ and $D_2$, one can require that the ratio between the cover in the two datasets is larger than a user defined threshold $\alpha$: $\frac{|cover(S,D_1)|}{|cover(S,D_2)|} \ge \alpha$. Let $D = D_1 \cup D_2$ and let $t_1 = \{ i | T_i \in D_1\}$ and $t_2 = \{ i | T_i \in D_2\}$ then we can extract the  discriminant patterns from $D$ by applying the following constraint. 
  \begin{equation}
    \label{eq:discr}
    \begin{gathered}
      \frac{\sum_{i \in t_1} {\var C_i}}{\sum_{i \in t_2} {\var C_i}}  \ge \alpha
    \end{gathered}
  \end{equation}
Such a constraint can also be used as an optimisation criterion in a CP framework.

  Type~4 constraints a.k.a. preference relations have been proposed in~\cite{dominance_dp} to formalise well-known pattern mining settings such as $maximal$ or $closed$ patterns. 
  Such preference relations can be enforced dynamically during search for any CP formulation~\cite{dominance_dp}. The preference relation for closed is $S' \succ S \equiv S \sqsubset S' \wedge cover(S,D) = cover(S',D)$ and one can reuse the global reified \textit{exists-embedding} constraint for this.

Finally, type~3 constraints over the inclusion relation are not possible in this model. Indeed, a new global constraint would have to be created for every possible (combination of) type~3 constraints. For example for \maxgap, one would have to modify Algorithm~\ref{alg:global_emp} to check whether the gap is smaller than the threshold, and if not, to search for an alternative embedding instead (thereby changing the complexity of the algorithm). 

\section{Decomposition with explicit embedding variables} \label{sec:second-model}

In the previous model, we used a global constraint to assign the ${\var C_i}$ variables to their appropriate value, that is:
${\var C_i} \leftrightarrow \exists e\mbox{ s.t. }{\var S} \sqsubseteq_e T_i$.  The global constraint efficiently tests the existence of one embedding, but does not expose the value of this embedding, thus it is impossible to express constraints over embeddings such as the \maxgap constraint. 

To address this limitation, we extend the previous model with a set of {\em embedding} variables ${\var E_{i1},\ldots,E_{i|T_i|}}$ that will represent an embedding $e = (e_1, \ldots, e_{|T_i|})$  of sequence ${\var S}$ in transaction $T_i$. In case there is no possible match for a character ${\var S_i}$ in $T_i$, the corresponding  ${\var E_{ij}}$ variable will be assigned a {\em no-match} value. 

\subsection{Variables and constraints}
 
\newcommand{\dummy}{{\em no-match~}}
\subsubsection{Embedding variables.}

\begin{figure}[t]
  \centering
  \Large
      \begin{tikzpicture}[scale=0.55, transform shape]
        \tikzstyle{seq_item}=[draw, fill=blue!20, shape=rectangle, minimum size=1cm];
        \tikzstyle{seq_item_small}=[draw, fill=blue!20, shape=rectangle, minimum size=1cm, minimum width=0.5cm];
        \tikzstyle{seq_item_static}=[draw, fill=white, shape=rectangle, minimum size=1cm];
        \tikzstyle{myedge}=[thick, -latex, color=red!80!black];

        \node at (0.3, 1) {${\var S}:~$}; 
        \setcounter{y}{1}
        \foreach \i in {A, B, $\epsilon$, $\epsilon$}{
          \node[seq_item, label=above:{\small p=\arabic{y}}] at (\arabic{y}, 0.3) (j\arabic{y}) {\i};
          \stepcounter{y}
        }

        
        \node[seq_item_small] (t1) at (-2.0, -1.4) {$1$};
        \node at (-3, -1.4) {${\var C_1}:~$};
                
        \node at (0.15, -1.4) {$T_1:~$}; 
        \setcounter{y}{1}
        \foreach \i in {A, C, B}{
          \node[seq_item_static] at (0+\arabic{y}, -1.4) (x1\arabic{y}) {\i};
          \stepcounter{y}
        }

        \node at (6.0, -1.4) {${\var E_1}:~$}; 
        \setcounter{y}{1}
        \foreach \i in {1, 3, {\it 4}, {\it 4}}{
          \node[seq_item, label=above:{\small j=\arabic{y}}] at (6+\arabic{y}, -1.4) (e1\arabic{y}) {\i};
          \stepcounter{y}
        }

        \draw (e11) edge[myedge, bend right, in=-130] (x11);
        \draw (e12) edge[myedge, bend right, out=300] (x13);


        \node[seq_item_small] (t2) at (-2.0, -2.4) {$0$};
        \node at (-3, -2.4) {${\var C_2}:~$};
                
        \node at (0.15, -2.4) {$T_2:~$}; 
        \setcounter{y}{1}
        \foreach \i in {B, A, A, C}{
          \node[seq_item_static] at (0+\arabic{y}, -2.4) (x2\arabic{y}) {\i};
          \stepcounter{y}
        }

        \node at (6.0, -2.4) {${\var E_2}:~$}; 
        \setcounter{y}{1}
        \foreach \i in {2, {\it 5}, {\it 5}, {\it 5}}{
          \node[seq_item] at (6+\arabic{y}, -2.4) (e2\arabic{y}) {\i};
          \stepcounter{y}
        }

        \draw (e21) edge[myedge, bend left, out=35, in=125] (x22);

        \end{tikzpicture}
\caption{Example assignment; blue boxes represent variables, white boxes represent data. The cursive values in ${\var E_1}$ and ${\var E_2}$ represent the \dummy value for that transaction.}
\label{fig:principle2}  
\end{figure}
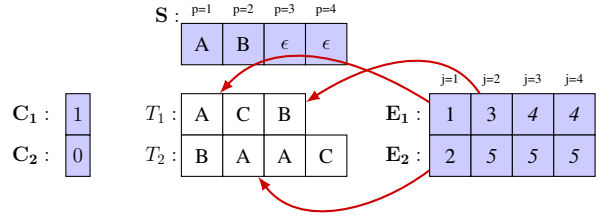

For each transaction $T_i$ of length $|T_i|$, we introduce integer variables ${\var E_{i1}}, \ldots, {\var E_{i|T_i|}}$. Each variable ${\var E_{ij}}$ is an index in $T_i$, and an assignment to ${\var E_{ij}}$ maps the variable ${\var S_j}$ to a position in  $T_i$; see Figure~\ref{fig:principle2}, the value of the index is materialized by the red arrows. The domain of ${\var E_{ij}}$ is initialized to all possible positions of $T_i$, namely $1, \ldots, |T_i|$ plus a \dummy entry which we represent by the value $|T_i| + 1$. 

\subsubspace
\subsubsection{The $\posmatch$ constraint.\label{par:posmatch}} This constraint ensures that the variables ${\var E_i}$ either represent an embedding $e$ such that ${\var S} \sqsubseteq_e T_i$ or otherwise at least one ${\var E_{ij}}$ has the \dummy value. Hence, each variable ${\var E_{ij}}$ is assigned the value $x$ only if the character in ${\var S_i}$ is equal to the character at position $x$ in $T_{i}$.  In addition, the constraint also ensures that the values between two consecutive variables ${\var E_{ij}, E_{i(j+1)}}$ are increasing so that the order of the characters in the sequence is preserved in the transaction. If there exist no possible match satisfying these constraints, the {\em no-match} value is assigned. 

\parspace
\begin{align}
  \label{eq:posmatch}
    \forall i \in 1, \ldots, n, \forall j \in 1,\ldots, |T_i|: &\quad ({\var S_j} = T_i[{\var E_{ij}}]) \lor ({\var E_{ij}} = |T_i|+1)\\
    \forall i \in 1, \ldots, n, \forall j \in 2,\ldots, |T_i|: &\quad ({\var E_{i(j-1)}} < {\var E_{ij}}) \lor ({\var E_{ij}} = |T_i|+1)
\end{align}

Here ${\var S_j} = T_{i}[{\var E_{ij}}]$ means that the symbol of ${\var S_j}$ equals the symbol at index ${\var E_{ij}}$ in transaction $T_i$. See Appendix~\ref{app:decomp} for an effective reformulation of these constraints.


\subsubspace
\subsubsection{\Isemb constraint.} Finally, this constraint ensures that a variable ${\var C_i}$ is $true$ if the embedding variables ${\var E_{i1},\ldots,E_{i|T_i|}}$ together  form a valid embedding of sequence ${\var S}$ in transaction $T_i$. More precisely: if each character ${\var S_j} \neq \epsilon$ is mapped to a position in the transaction that is different from the \dummy value. 
\begin{equation}
  \label{eq:isemb}
    \forall i \in 1, \ldots, n:\quad 
    {\var C_i} \leftrightarrow \forall j \in 1,\ldots, |T_i|: ~ ({\var S_j} \neq \epsilon) \rightarrow ({\var E_{ij}} \neq |T_i|+1)
\end{equation}
\noindent Note that depending on how the ${\var E_{ij}}$ variables will be searched over, the above constraints are or are not equivalent to enforcing ${\var C_i} \leftrightarrow \exists e\mbox{ s.t. }{\var S} \sqsubseteq_e T_i$. This is explained in the following section.

\subsection{Search strategies for checking the existence of embeddings}

CP's standard enumerative search would search for all satisfying assignments to the ${\var S_j}, {\var C_i}$ and ${\var E_{ij}}$ variables. As for each sequence of size $m$, the number of embeddings in a transaction of size $n$ can be  $O(n^m)$, such a search would not perform well. Instead, we only need to search whether {\em one} embedding exists for each transaction.


\optional{
\subsubspace
\subsubsection{Without additional constraints on ${\var E_{ij}}$ and ${\var C_i}$.}
The reformulation in Appendix \ref{app:decomp} of the \posmatch constraint is \textit{lower}-bound consistent on the ${\var E_{ij}}$ variables, assuming there are no other constraints on the ${\var E_{ij}}$ or ${\var C_i}$ variables. Indeed, for any assignment to ${\var S}$, taking the smallest value of each ${\var E_{ij}}$ variable results in an assignment that is a valid embedding for all transactions that admit one. Thus in that specific case, there is no need to search over the ${\var E_{ij}}$ or ${\var C_i}$ variables. Indeed, after searching over the ${\var S_j}$ variables, either ${\var C_i}$ is {\em false} or it is unassigned and can be set to {\em true} by assigning each ${\var E_{ij}}$ to the lowest value in its domain (non \dummy value, otherwise ${\var C_i}$ would be \textit{false}).
}

\subsubspace
\subsubsection{With additional constraints on ${\var E_{ij}}$ but not ${\var C_i}$.}
When there are additional constraints on the ${\var E_{ij}}$ variables 
such as \maxgap, one has to perform backtracking search to find a valid embedding. We do this after the ${\var S}$ variables have been assigned.

We call the search over the ${\var S}$ variables the \textit{normal} search, and the search over the ${\var E_{ij}}$ variables the \textit{sub} search. Observe that one can do the \textit{sub} search for each transaction $i$ independently of the other transactions as the different ${\var E_{i}}$ have no influence on each other, only on ${\var C_i}$. Hence, one does not need to backtrack across different \textit{sub} searchers.

The goal of a \textit{sub} search for transaction $i$ is to find a valid embedding for that transaction. Hence, that \textit{sub} search should search for an assignment to the ${\var E_{ij}}$ variables with ${\var C_i}$ set to \textit{true} first. If a valid assignment is found, an embedding for $T_i$ exists and the \textit{sub} search can stop. If no assignment is found, ${\var C_i}$ is set to false and the \textit{sub} search can stop too.
See Appendix \ref{app:subsearch}
for more details on the sub search implementation.

\subsubspace
\subsubsection{With arbitrary constraints.} The constraint formulation in Equation \eqref{eq:isemb} is not equivalent to ${\var C_i} \leftrightarrow \exists e\mbox{ s.t. }{\var S} \sqsubseteq_e T_i$. For example, lets say some arbitrary constraint propagates ${\var C_i}$ to \textit{false}. For the latter constraint, this would mean that it will enforce that ${\var S}$ is such that there does not exists an embedding of it in $T_i$. In contrast, the constraint in Equation \eqref{eq:isemb} will propagate some ${\var E_{ij}}$ to the \dummy value, even if there exists a valid match for the respective ${\var S_j}$ in $T_i$!

To avoid an ${\var E_{ij}}$ being set to the \dummy value because of an assignment to ${\var C_i}$, we can replace Equation \eqref{eq:isemb} by the half-reified $\forall i: ~{\var C_i} \rightarrow ( \forall j ~({\var S_j} \neq \epsilon) \rightarrow ({\var E_{ij}} \neq |T_i|+1)~)$ during \textit{normal} search.

The \textit{sub} search then has to search for a valid embedding, even if ${\var C_i}$ is set to \textit{false} by some other constraint.
One can do this in the \textit{sub} search of a specific transaction $i$ by replacing the respective half-reified constraint by the constraint $~{\var C'_i} \leftrightarrow ( \forall j ~({\var S_j} \neq \epsilon) \rightarrow ({\var E_{ij}} \neq |T_i|+1)~)$ over a new variable ${\var C'_i}$ that is local to this \textit{sub} search. The \textit{sub} search can then proceed as described above, by setting ${\var C'_i}$ to \textit{true} and searching for a valid assignment to ${\var E_{i}}$. Consistency between ${\var C'_i}$ and the original ${\var C_i}$ must only be checked after the \textit{sub} search for transaction $i$ is finished. This guarantees that for any solution found, if ${\var C_i}$ is \textit{false} and so is ${\var C'_i}$ then indeed, there exists no embedding of ${\var S}$ in $T_i$.

\subsection{Projected frequency}
Each ${\var E_{ij}}$ variable represents the positions in $T_i$ that ${\var S_j}$ can still take. This is more general than the projected transaction, as it also applies when the previous symbol in the sequence ${\var S_{j-1}}$ is not assigned yet. 
Thus, we can also use the ${\var E_{ij}}$ variables to require that every symbol of ${\var S_j}$ must be frequent in the (generalised) projected database. This is achieved as follows. 
\begin{equation}
    \label{eq:freq-char-reif}
    \begin{gathered}
      \forall j \in 1\ldots n, \forall x \in \Sigma,
      {\var S_j} = x \rightarrow |\{ i : {\var C_i} \wedge T_i[{\var E_{ij}}] = x \}| \ge \theta
    \end{gathered}
\end{equation}

\noindent See Appendix~\ref{app:projfreq}
for a more effective reformulation.

\optional{
Even when reformulated, this is a costly constraint which propagates to all ${\var S_j}$, independent of the current prefix. An alternative is to devise another specialised search routine that checks the frequency over all $T_i$ and ${\var E_{ij}}$, just before branching over an ${\var S_j}$.
}

\subsection{Constraints}
\label{sec:user-constraints}
All constraints from Section~\ref{sec:user-constraints1} are supported in this model too. Additionally, constraints over the inclusion relations are also supported; for example, \maxgap and \maxspan. Recall from Section~\ref{sec:constraints-max-1} that for an embedding  $e = (e_1, \ldots, e_k)$, we have $\maxgap_i(e)  \Leftrightarrow \forall j \in 2\ldots |T_i|, (e_{j} - e_{j-1} - 1) \le \gamma$. One can constrain all the embeddings to satisfy the \maxgap constraint as follows  (note how $x$ is smaller than the \dummy value $|T_i|+1$):
\begin{align}
  \label{eq:maxgap}
  \forall i \in 1\ldots n, \forall j \in 2\ldots |T_i|, x \in 1\ldots |T_i|:
  \quad {\var E_{ij}} = x \rightarrow x - {\var E_{i(j-1)}} \le \gamma+1
\end{align}
\Maxspan was formalized as $\maxspan_i(e) \Leftrightarrow e_{|T_i|} - e_1 + 1 \le \gamma$ and can be formulated as a constraint as follows:
\begin{align}
  \label{eq:maxspan}
  \forall i \in 1\ldots n, \forall j \in 2\ldots |T_i|, x \in 1\ldots |T_i|:
  \quad {\var E_{ij}} = x \rightarrow x - {\var E}_{{\bf i}1} \le \gamma-1
\end{align}
In practice, we implemented a simple \textit{difference-except-no-match} constraint that achieves the same without having to post a constraint for each $x$ separately.

\section{Experiments}
\label{sec:experiments}
\optional{
  \todo{adapt}
  \renewcommand{\r}[1]{\parbox[t]{2mm}{{\rotatebox[origin=l]{60}{#1}}}}
  \newcommand{\cm}{\checkmark}
  \begin{table*}
    \centering\small
    \begin{tabular}{|c||cccc|cccc|cccc|} 
      \hline
      task $\rightarrow$ & \multicolumn{4}{c}{frequent} & \multicolumn{4}{c}{closed} & \multicolumn{4}{c|}{relevant}                                          \\\hline
      solver $\downarrow$          & none                         & gap                        & span & both & none  & gap   & span  & both  & none & gap & span & both \\\hline
      cSpade    & \cm                          & \cm                        & \cm  &      &       &       &       &       &      &     &      &      \\\hline
      Bide      &                              &                            &      &      & \cm   &    &       &       &      &     &      &      \\\hline
      CP-SM     & \cm                          & \cm                        & \cm  & \cm  & \cm & \cm & \cm & \cm &      &     &      &      \\\hline
      Bide+P.P.  &                              &                            &      &      &       &       &       &       & \cm  &     &      &      \\\hline
      {\bf RCP}  & \cm                          & \cm                        & \cm  & \cm  & \cm   & \cm   & \cm   & \cm   & \cm  & \cm & \cm  & \cm  \\\hline
    \end{tabular}

    \caption{Capabilities of various solvers.}
    \label{tab:cando}
  \end{table*}
}
The goal of these experiments is to answer the four following questions:
{\bf Q1:} What is the overhead of exposing the embedding variables in the {\em decomposed} model?
{\bf Q2:} What is the impact of using projected frequency in our models?
{\bf Q3:} What is the impact of adding constraints on runtime and on number of results? 
{\bf Q4:} How does our approach compares to existing methods? 

\parspace
\paragraph{Algorithm and execution environment:}
All the models described in this paper have been implemented in the Gecode solver\footnote{http://www.gecode.org}. We compare our {\em global} and {\em decomposed} models  (Section~\ref{sec:first-model} and Section~\ref{sec:second-model}) to the state-of-the-art algorithms cSpade\cite{zaki2000sequence} and PrefixSpan~\cite{han2001prefixspan}. We use the author's cSpade implementation\footnote{http://www.cs.rpi.edu/~zaki/www-new/pmwiki.php/Software/} and a publicly available PrefixSpan implementation by Y. Tabei\footnote{https://code.google.com/p/prefixspan/}.  We also compare our models to the CP-based approach proposed by \cite{metivierconstraint}. No implementation of this is available so we reimplemented it in Gecode. Gecode does not support non-deterministic automata so we use a more compact DFA encoding that requires only $O(n*|\Sigma|)$ transitions, by constructing it back-to-front. We call this approach {\em regular-dfa}. Unlike the non-deterministic version, this 
does not allow the addition of constraints of type~3 such as \maxgap.

All algorithms were run on a Linux PC with 16~GB of memory. Algorithm runs taking more than 1 hour or 
more than 75\% of the RAM were terminated. The implementation and the datasets used for the experiments are available online \footnote{https://dtai.cs.kuleuven.be/CP4IM/cpsm}. 

\parspace
\paragraph{Datasets:}
The datasets used are from real data and have been chosen to represent a variety of application domains.
In {\bf Unix user}\footnote{https://archive.ics.uci.edu/ml/datasets/}, each transaction is a series of shell commands executed by a user during one session. We report results on User~3; results are similar for the other users. 
%
\noindent{\bf JMLR} is a natural language processing dataset; each transaction is an abstract of a paper from the {\em Journal of Machine Learning Research}. 
\noindent{\bf iPRG} is a proteomics dataset from the application described in \cite{trypticcleavage}; each transaction is a sequence of peptides that is known to cleave in presence of a Trypsin enzyme.
{\bf FIFA} is click stream dataset\footnote{{http://www.philippe-fournier-viger.com/spmf/}} from logs of the website of the FIFA world cup in 98; each transaction is a sequence of webpages visited by a user during a single session.  Detailed characteristics of the datasets are given in Table~\ref{tab:dataset-spec}. Remark that the characteristic of these datasets are very diverse due to their different origins.

In our experiments, we vary the minimum frequency threshold ($minsup$). Lower values for $minsup$ result in larger solution sets, thus in larger execution times.

\begin{table}[t]
  \centering
  {\footnotesize
\begin{tabular}{|l|c|c|c|c|c|c|}
\hline
dataset    & $|\Sigma|$       & $|\D|$     & $||\D||$                   & $\displaystyle\max_{ T \in \D} |T|$ & $\mbox{avg}~|T|$ & density \\\hline 
Unix user    & 265          & 484        & 10935                      & 1256                                       & 22.59                                            & 0.085    \\\hline 
JMLR  & 3847         & 788      & 75646                      & 231                                        & 96.00                                             & 0.025 \\\hline
iPRG  & 21         & 7573      & 98163                      & 13                                        & 12.96                                             & 0.617 \\\hline
FIFA & 20450 & 2990 & 741092& 100 & 36.239 & 0.012 \\\hline
\end{tabular}}
  \caption{Dataset characteristics. Respectively: dataset name, number of distinct symbols, number of transactions, total number of symbols in the dataset, maximum transaction length, average transaction length, and density calculated by $\frac{||\D||}{|\Sigma| \times |\D|}$.}
  \label{tab:dataset-spec}
\end{table}

\begin{figure*}[h]
\centering 
\includegraphics[width=0.255\textwidth]{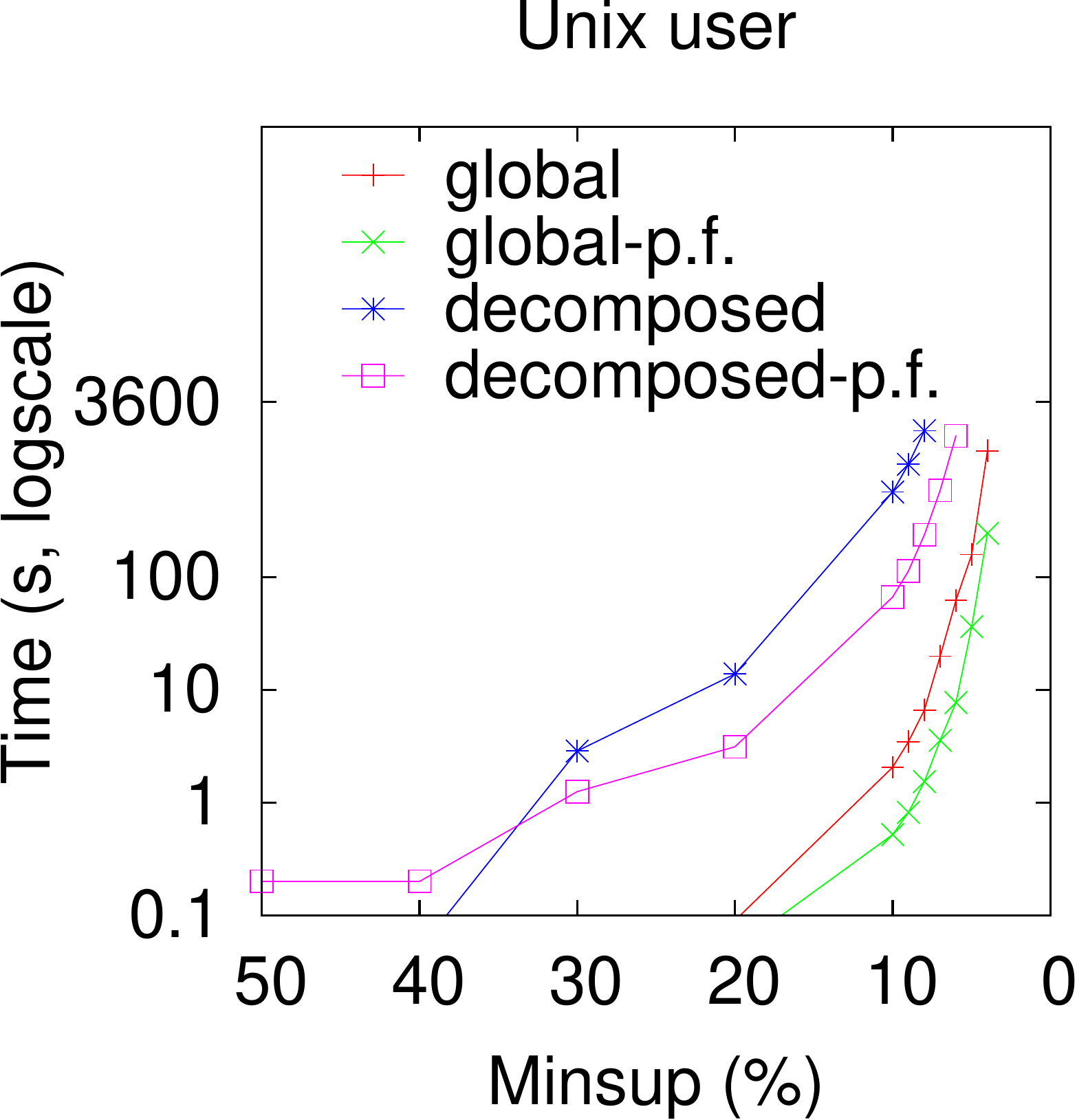}\hfill
    \includegraphics[width=0.24\textwidth]{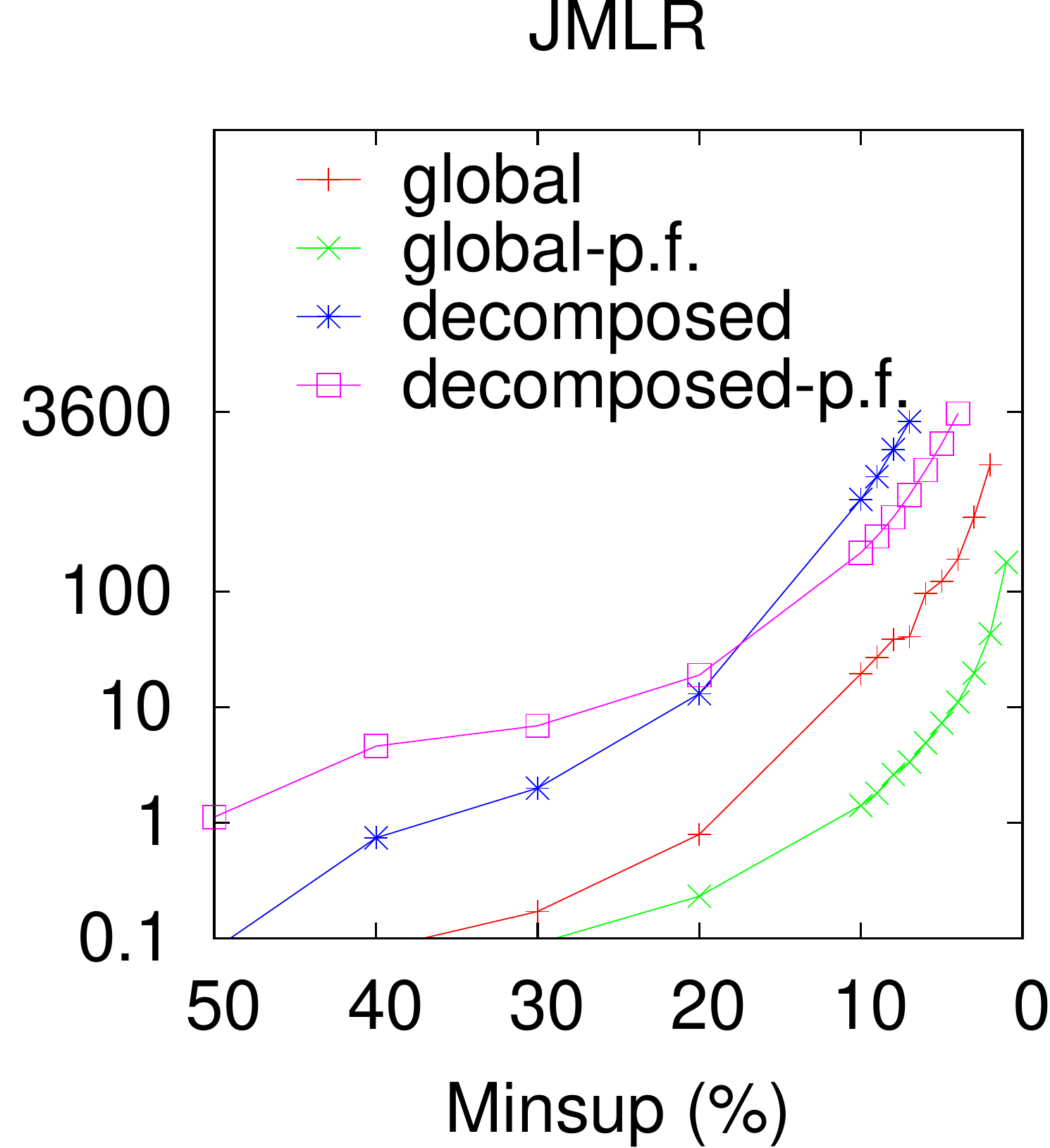}\hfill
\includegraphics[width=0.24\textwidth]{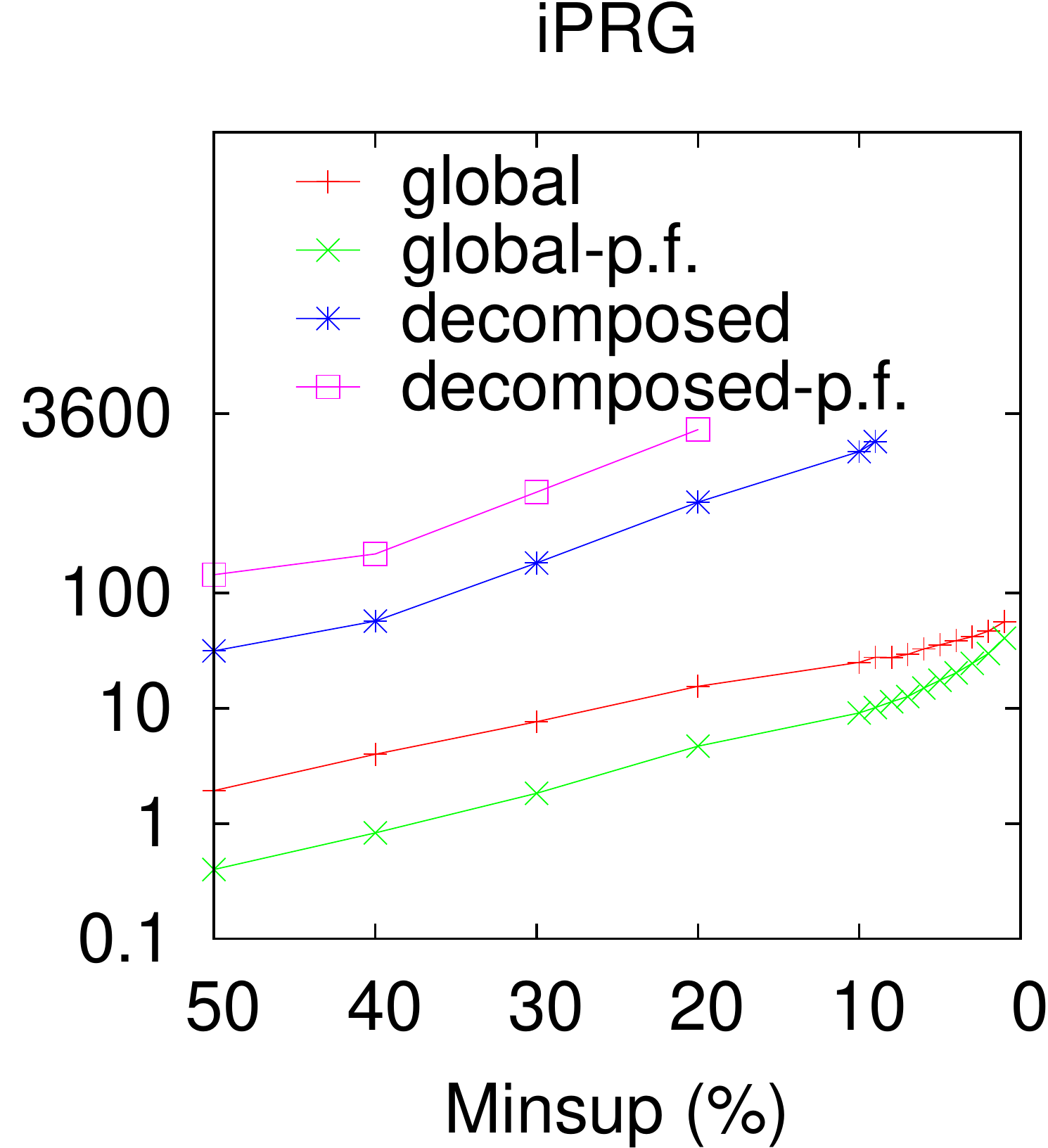}\hfill
\includegraphics[width=0.24\textwidth]{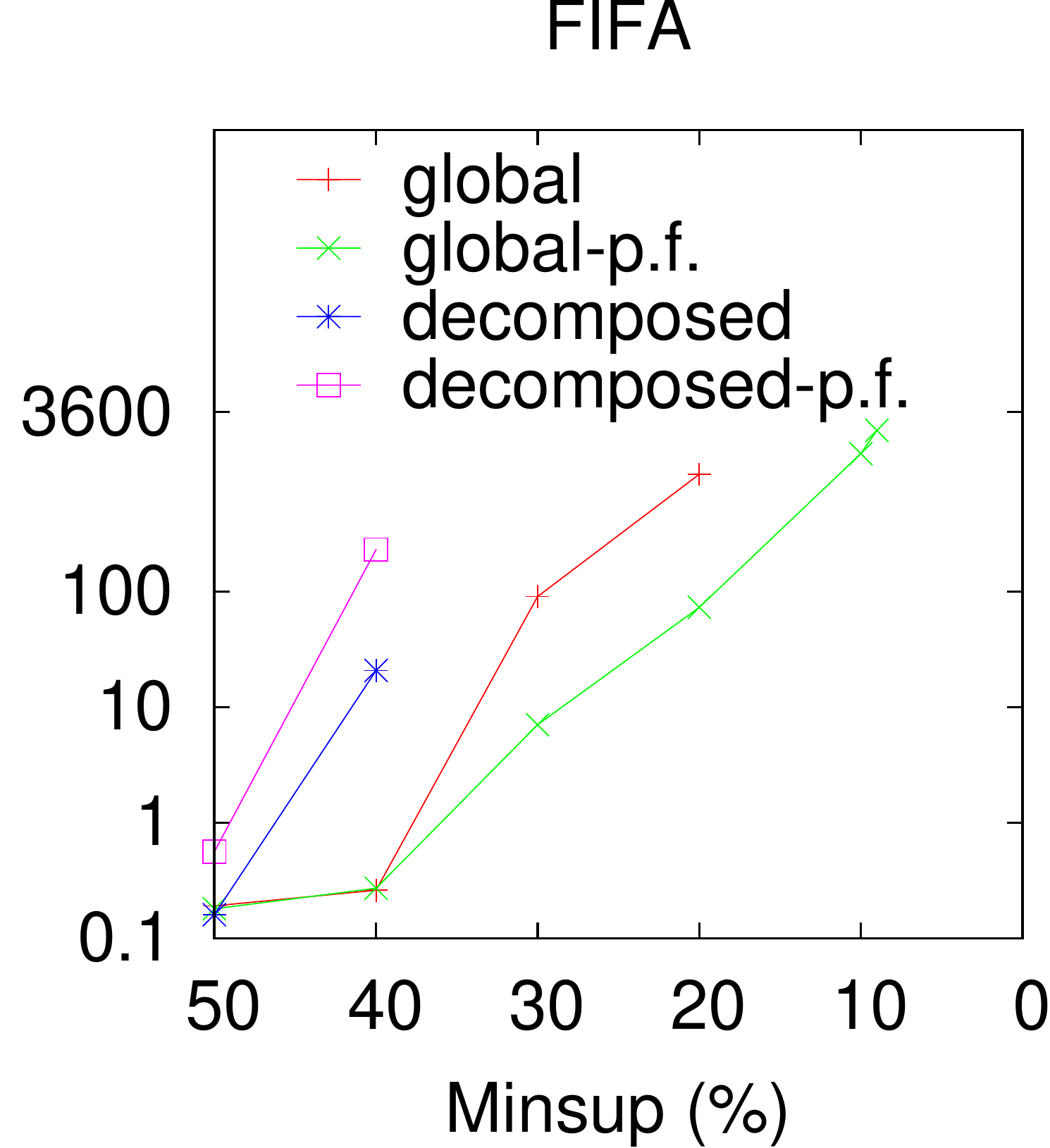}

 


  \caption{Global model vs. decomposed model: Execution times. (Timeout 1 hour.)}
\label{fig:time_all}
\end{figure*}

\parspace
\paragraph{Experiments:}
First we compare the {\em global} and the {\em decomposed} models. The execution times for these models are shown on Fig.~\ref{fig:time_all}, both without and with projected frequency (indicated by {\em -p.f.}). We first look at  the impact of exposing the embedding variables in the {\em decomposed} model ({\bf Q1}). Perhaps unsurprisingly, the {\em global} model is up to one order of magnitude faster than the {\em decomposed} model,  which has $O(n*k)$ extra variables.  This is the overhead required to allow one to add constraints over the inclusion relation. We also study the impact of the projected frequency on both models ({\bf Q2}). In the \textit{global} model this is done as part of the search, while in the \textit{decomposed} model this is achieved with an elaborate constraint formulation. For {\em global-p.f.} we always observe a speedup in Fig.~\ref{fig:time_all}. Not so for {\em decomposed-p.f.} for the two largest (in terms of $||D||$) datasets.

\begin{figure*}[t]
\centering 
  \includegraphics[width=0.243\textwidth]{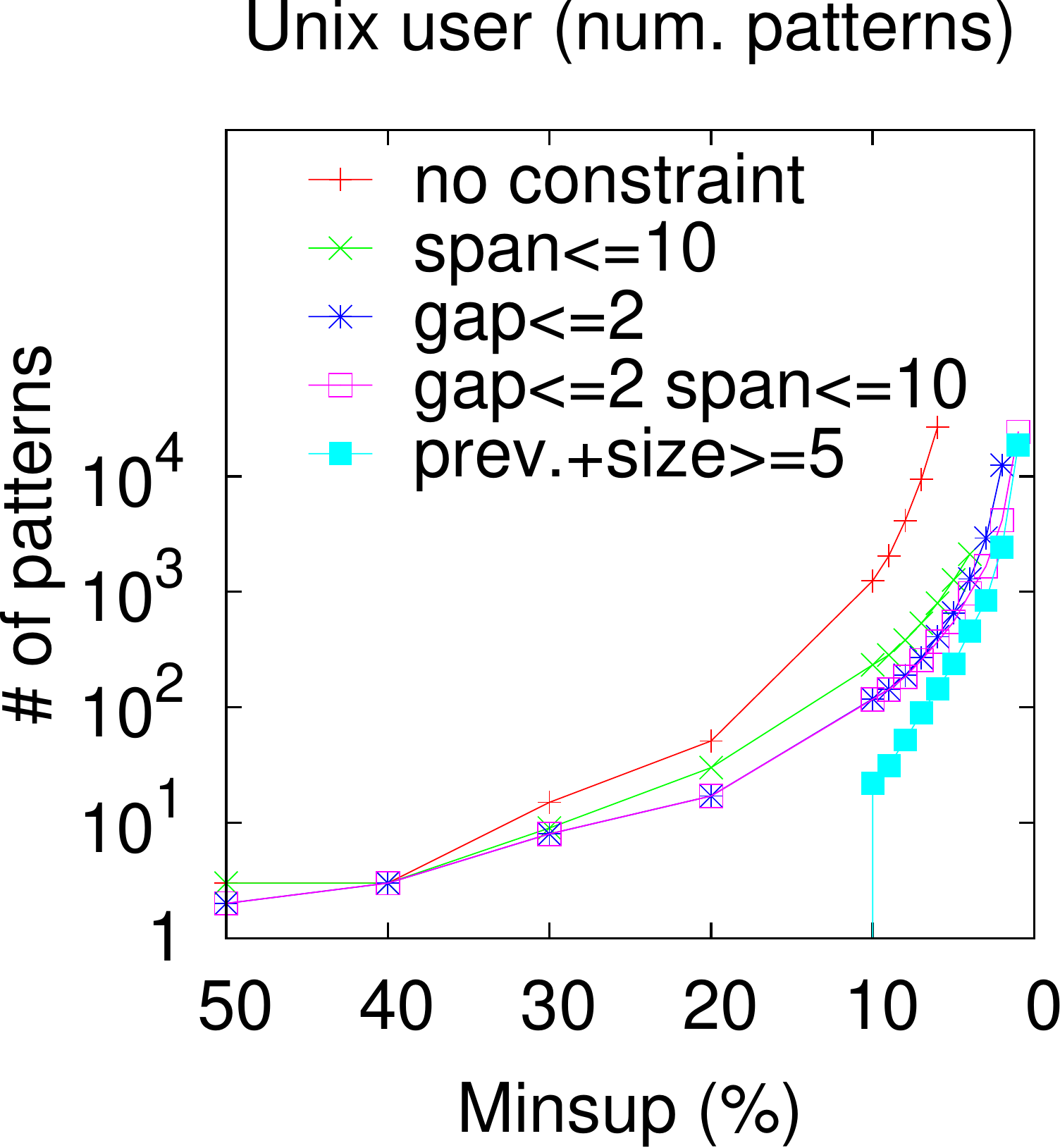}\hfill
    \includegraphics[width=0.227\textwidth]{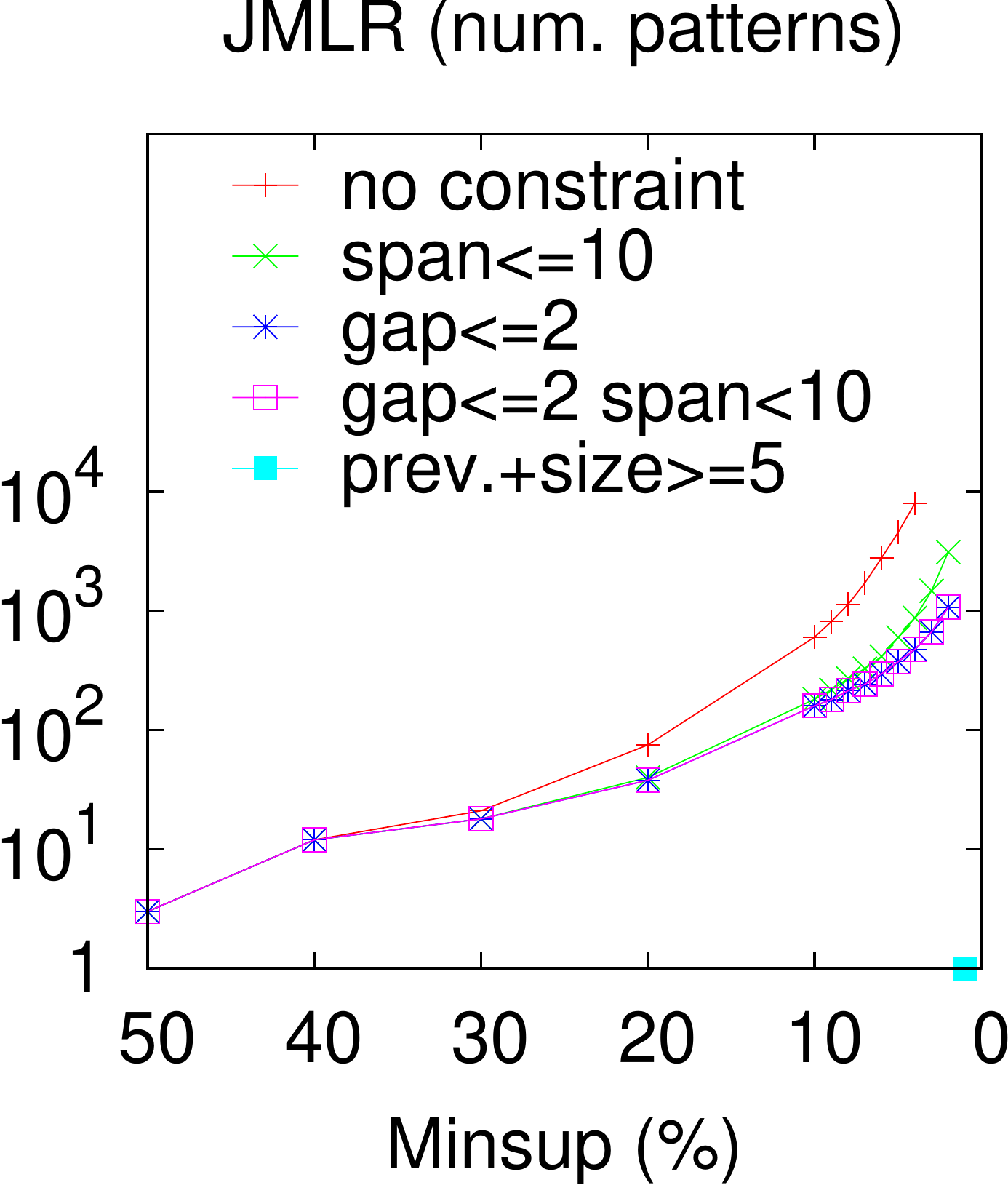}\hfill
\includegraphics[width=0.227\textwidth]{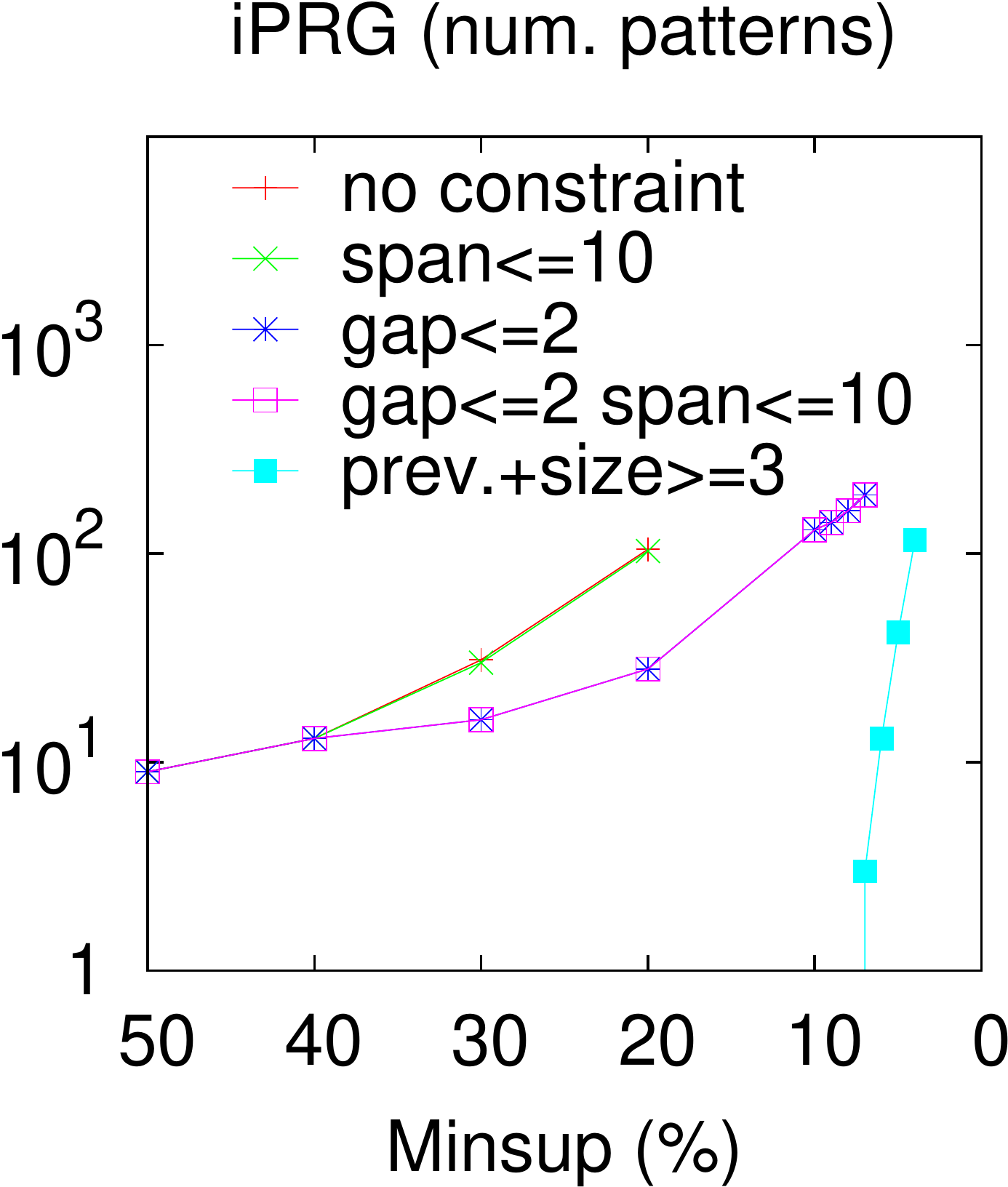}\hfill
\includegraphics[width=0.227\textwidth]{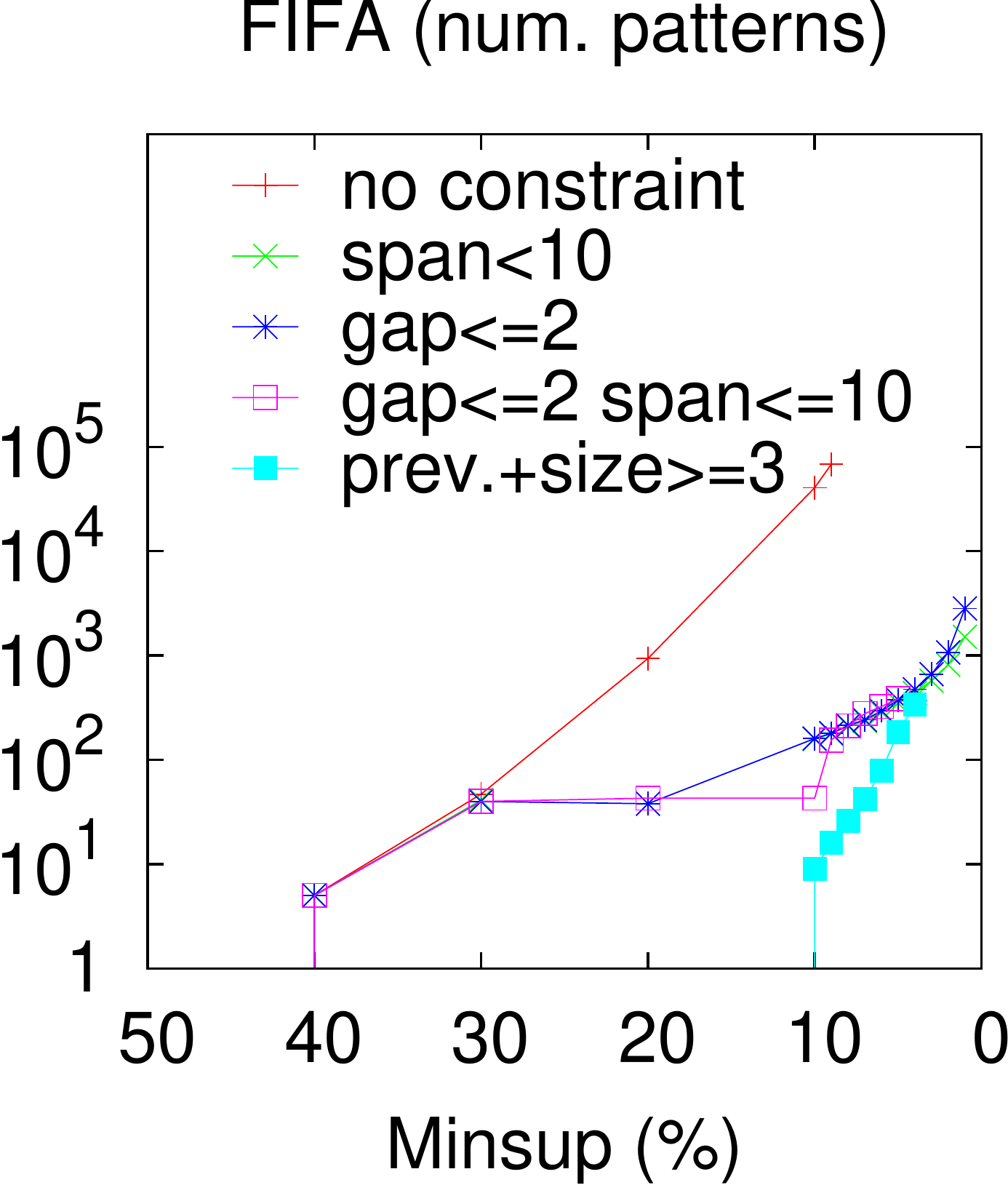}  \\
  \includegraphics[width=0.255\textwidth]{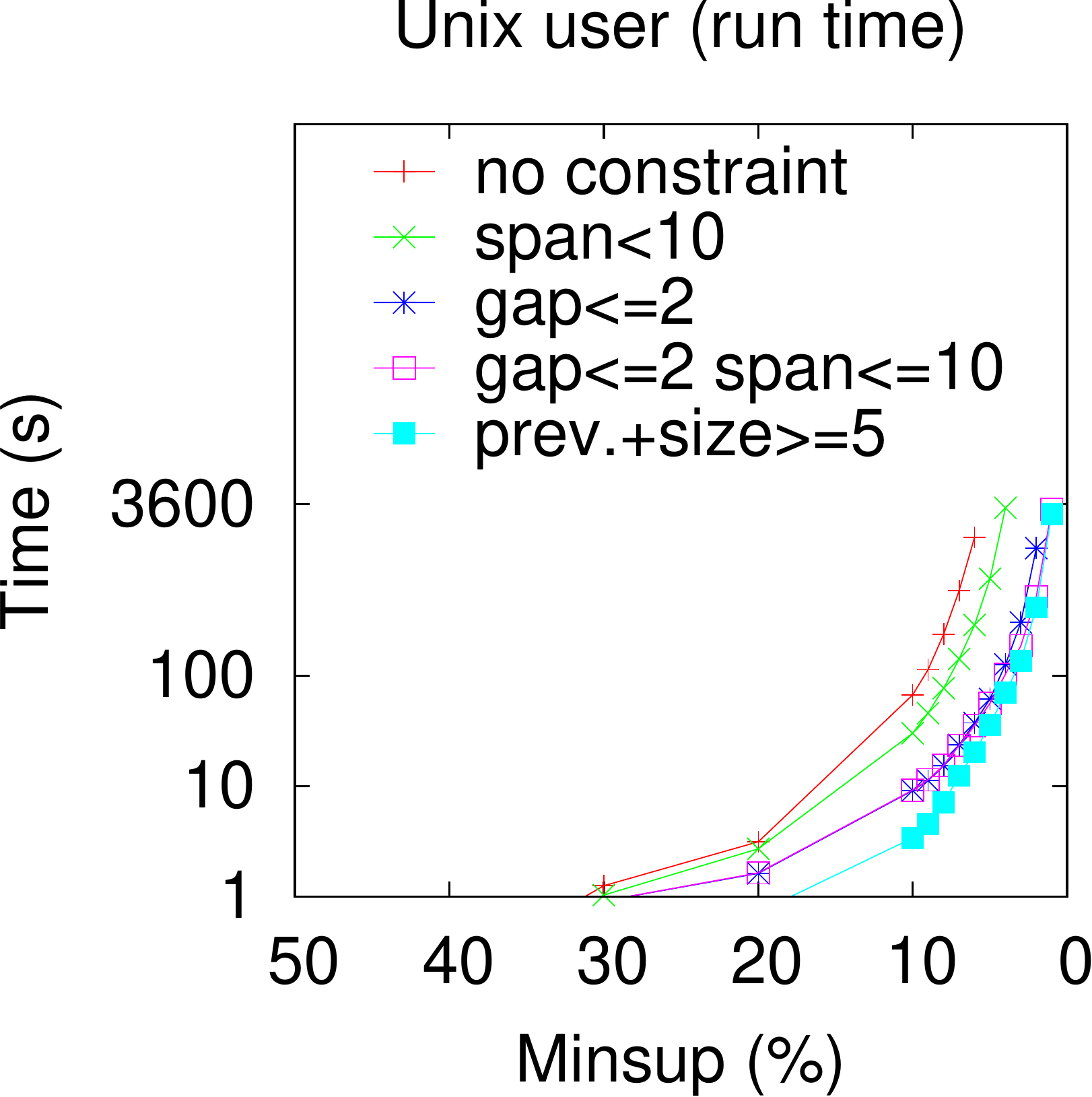}\hfill
    \includegraphics[width=0.24\textwidth]{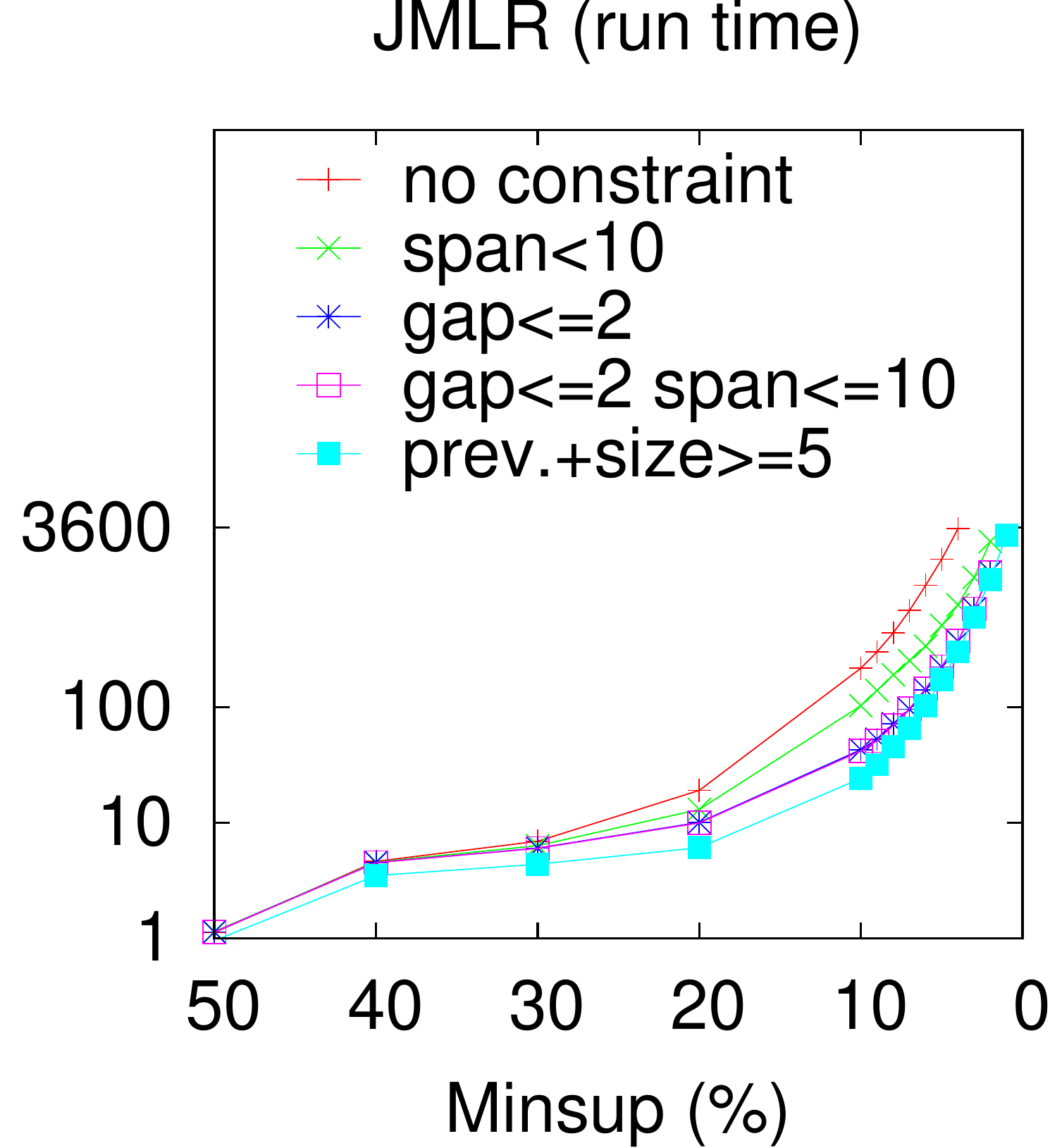}\hfill
    \includegraphics[width=0.24\textwidth]{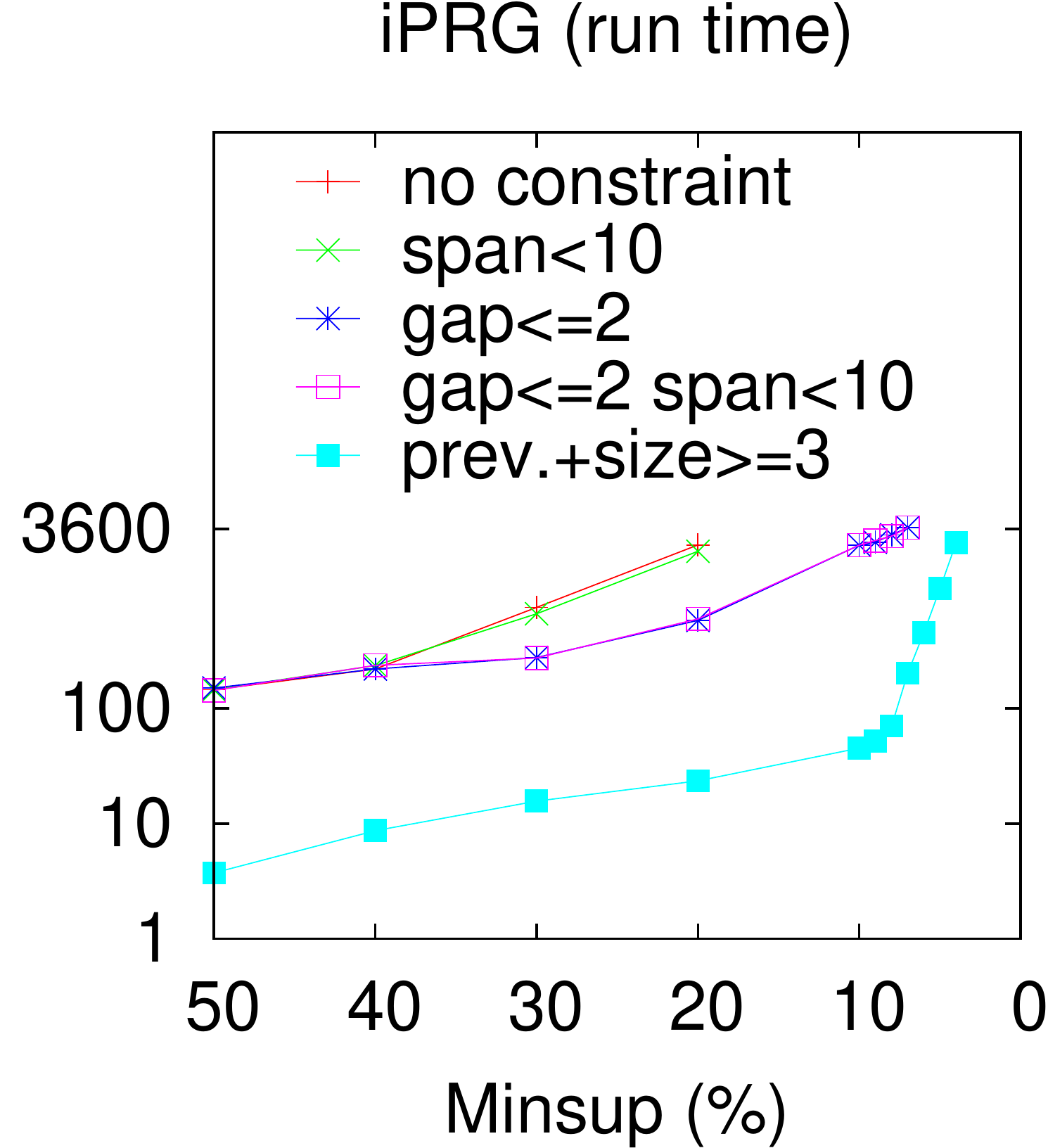}\hfill
   \includegraphics[width=0.24\textwidth]{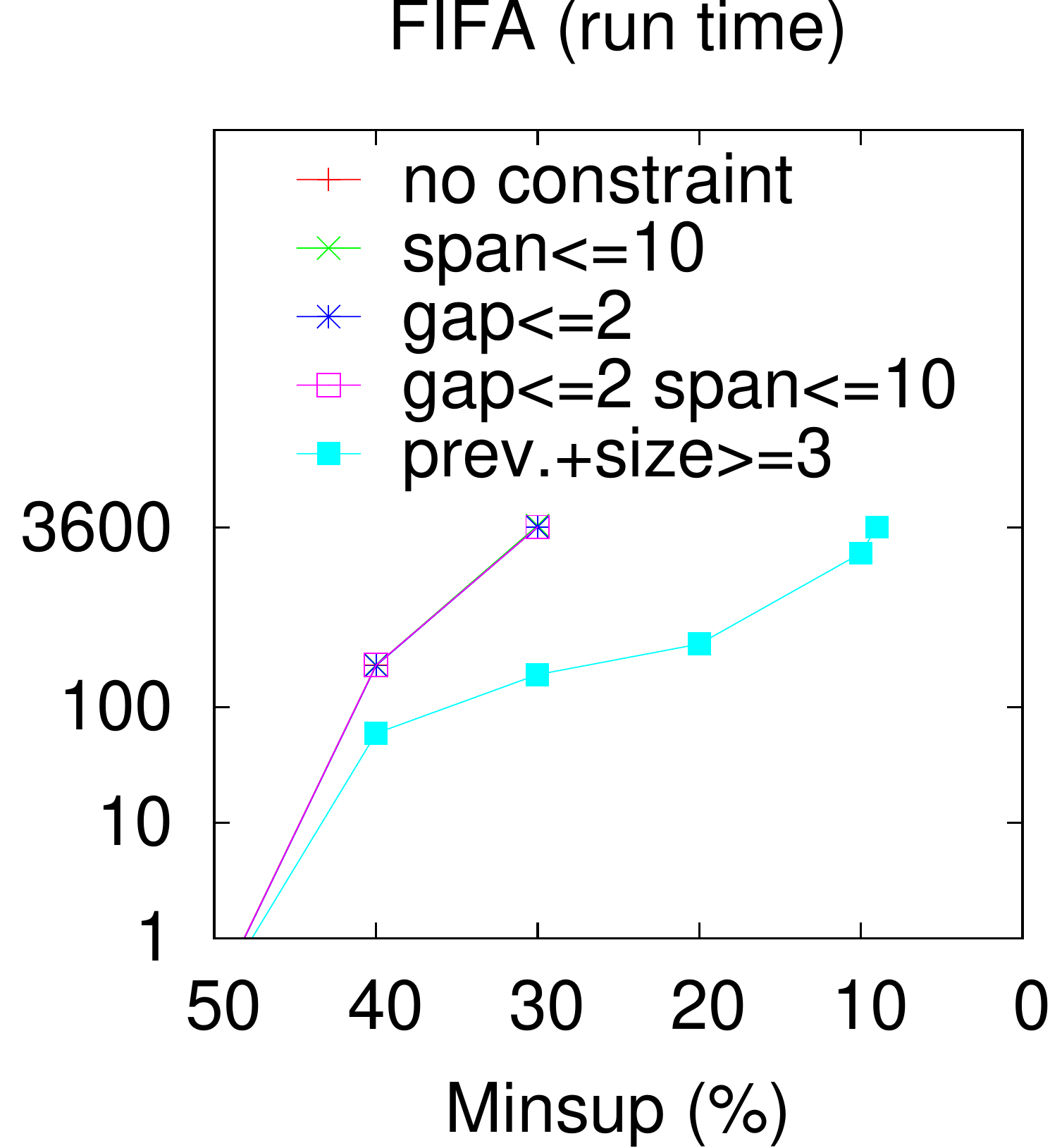} 
\caption{Number of patterns (top) and execution times (bottom) for the decomposed model with various combinations of constraints.}
\label{fig:numsol}
\end{figure*}

We now evaluate the impact of user constraints on the number of results and on the execution time ({\bf Q3}). Fig.~\ref{fig:numsol} shows the number of patterns and the execution times for various combinations of constraints.  We can see that adding constraints 
enables users to control the explosion of the number of patterns, and that the execution times decrease accordingly.  The constraint propagation 
allows early pruning of invalid solutions which  effectively compensates the computation time of checking the constraints. 
For example, on the Unix user dataset, it is not feasible to mine for patterns at 5\% minimum frequency without constraints, let alone do something with the millions of patterns found. 
On the other hand, by adding constraints one can look for interesting patterns at low frequency without being overwhelmed by the number of results (see also later).

\optional{By using combinations of relevant constraints, analysts can look for interesting patterns at low frequency without being overwhelmed by the number of results. } 

\begin{figure*}[t]
\centering 
\includegraphics[width=0.255\textwidth]{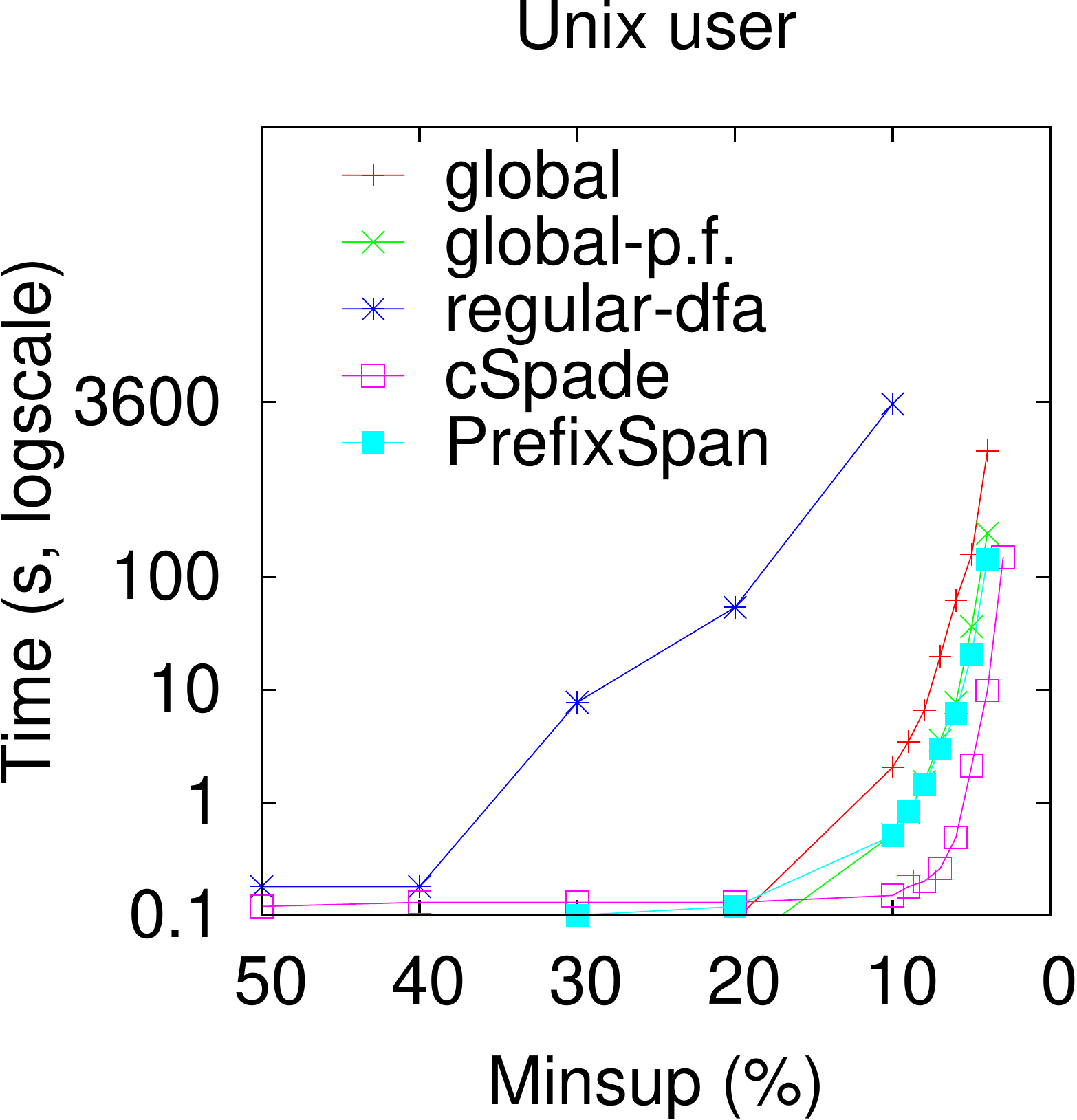}\hfill
    \includegraphics[width=0.24\textwidth]{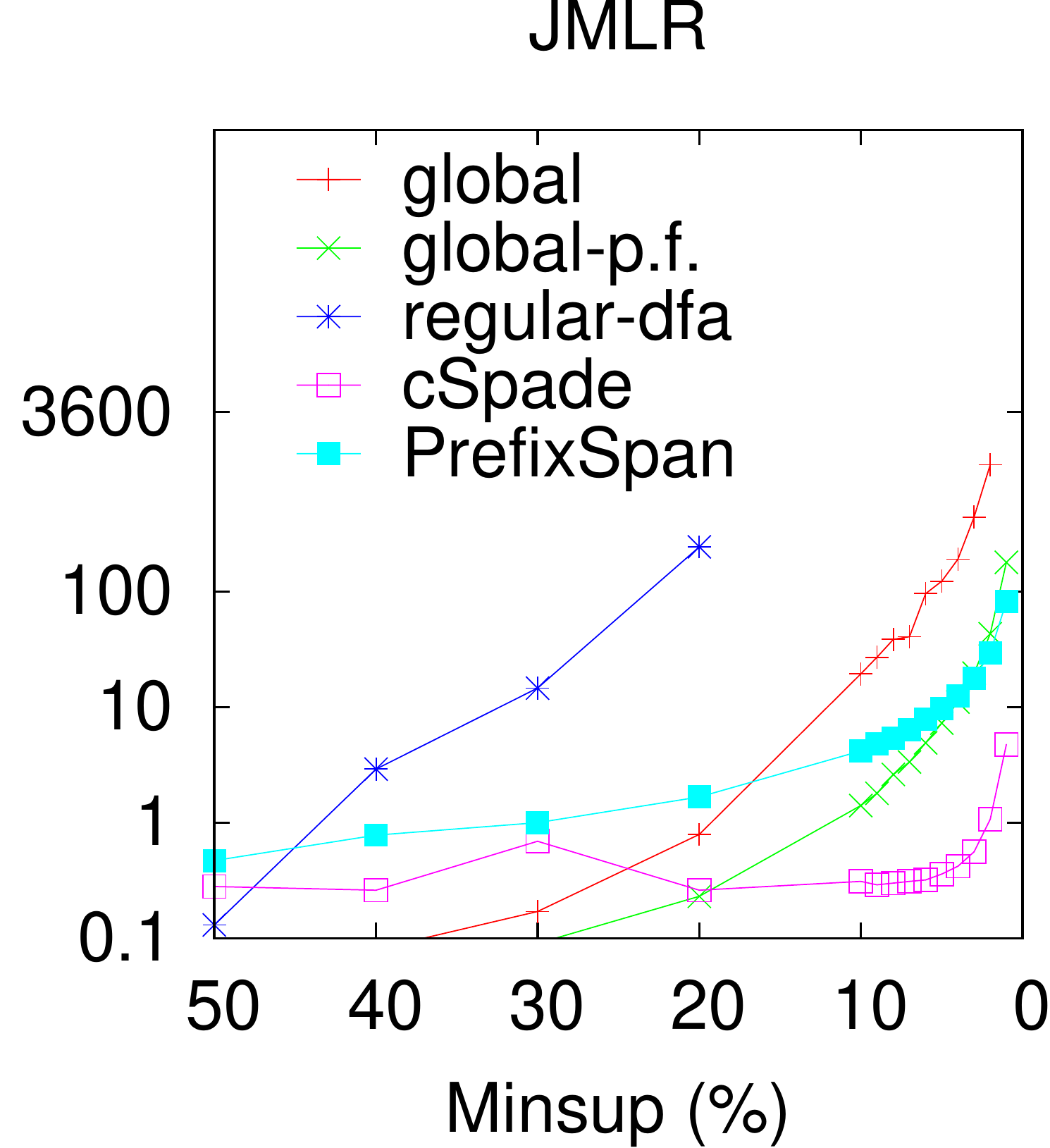}\hfill
    \includegraphics[width=0.24\textwidth]{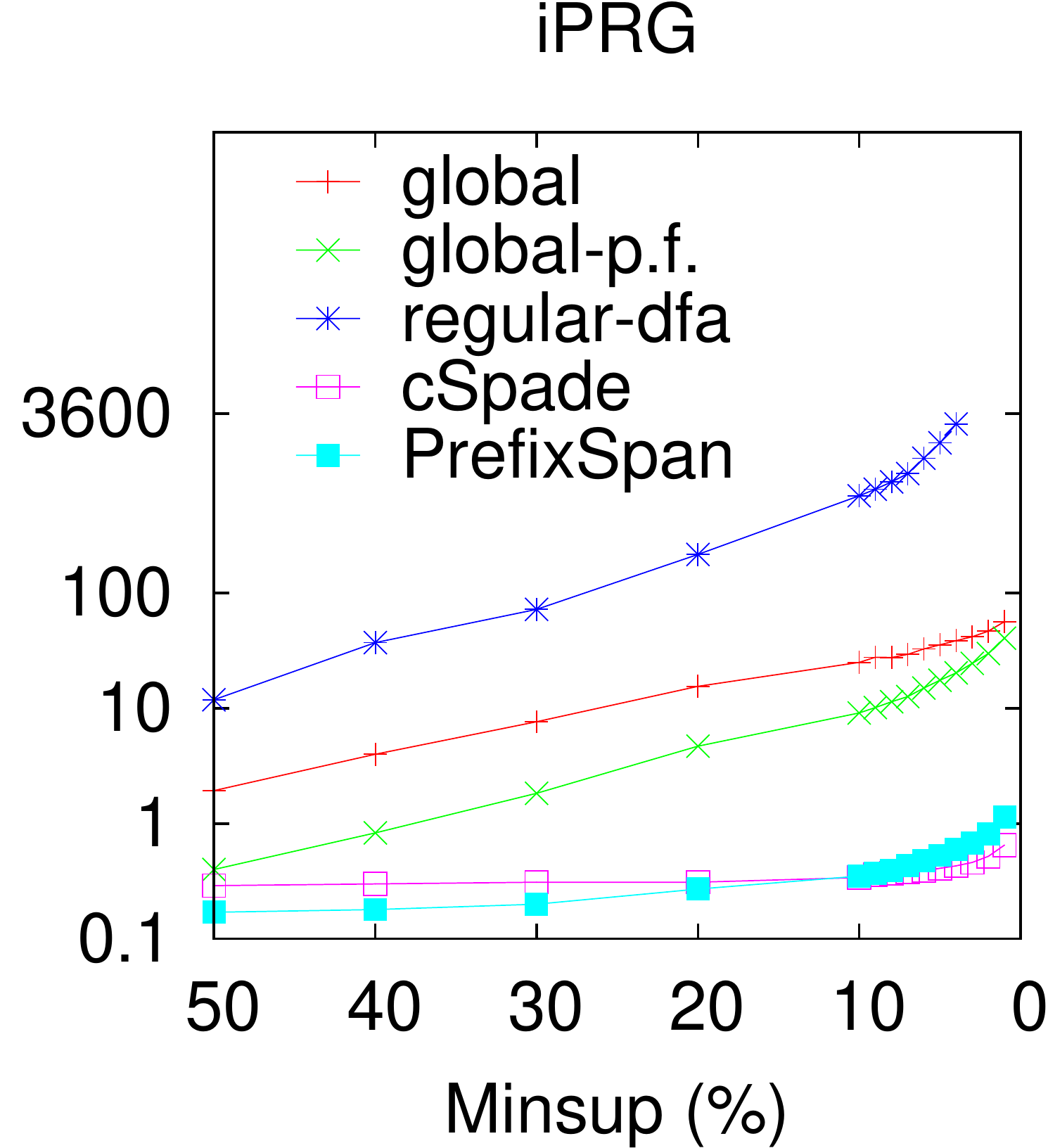}\hfill
    \includegraphics[width=0.24\textwidth]{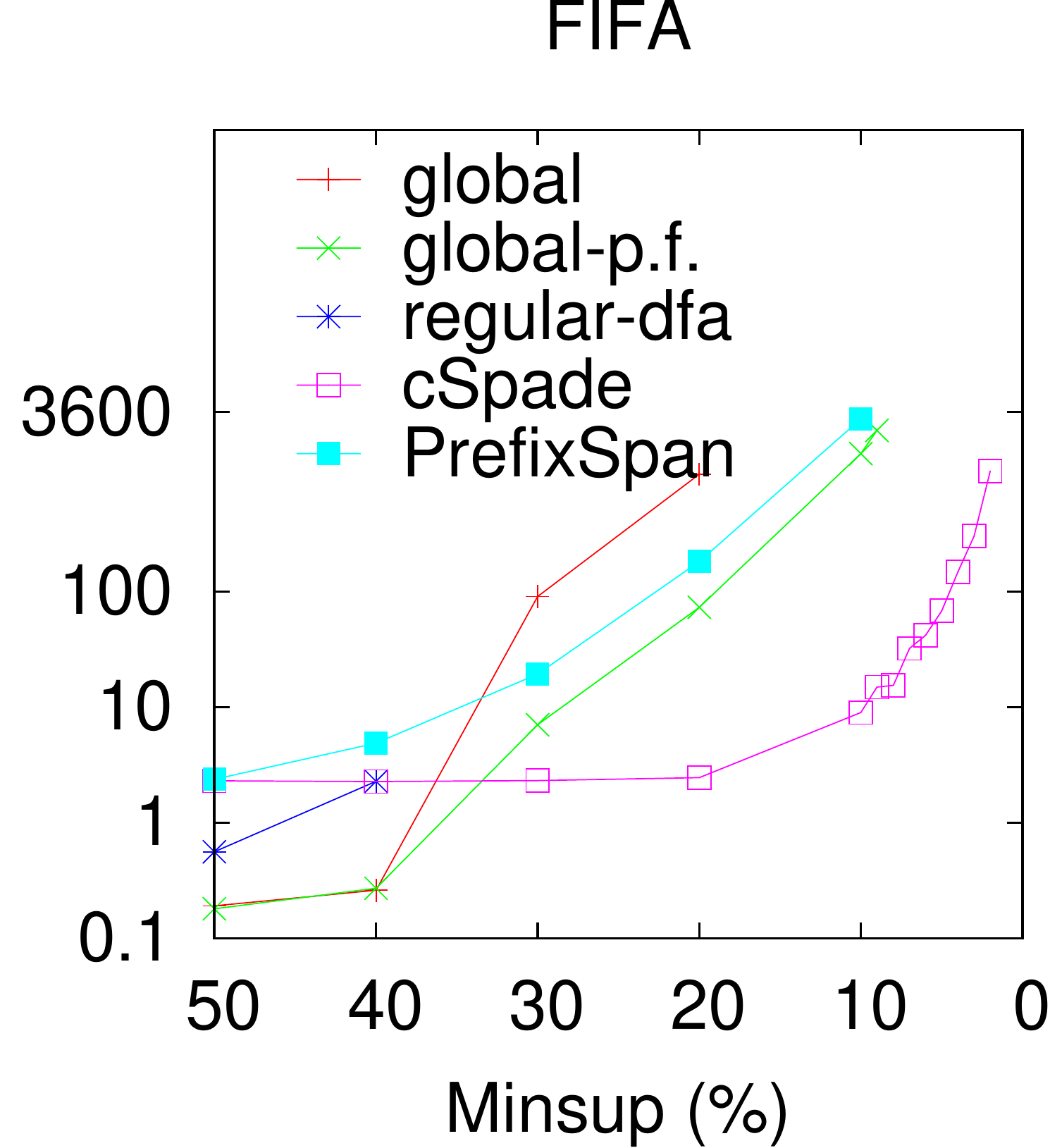}\\
  \caption{Global model vs. other approaches. Execution times. (Timeout 1 hour.)}
\label{fig:comparatives}
\end{figure*}

The last experiment compares our models to existing algorithms. Fig.~\ref{fig:comparatives} shows the execution times for our {\em global} model compared with {\em regular-dfa}, PrefixSpan and cSpade ({\bf Q4}). 
First, we can observe that {\em regular-dfa} is always slowest. On iPRG it performs reasonably well, but the number of transitions in the DFAs does not permit it to perform well on datasets with a large alphabet or large transactions, such as Unix user, JMLR or FIFA. Furthermore, it can not make use of projected frequencies.

\textit{global} shows similar, but much faster, behaviour than \textit{regular-dfa}. On datasets with many symbols such as JMLR and FIFA, we can see that not using projected frequency is a serious drawback; indeed, \textit{global-p.f.} performs much better than \textit{global} there.

Of the specialised algorithms, \textit{cSpade} performs better than \textit{PrefixSpan}; it is the most advanced algorithm and is the fastest in 
all experiments (not counting the highest frequency thresholds). \textit{global-p.f.} has taken inspiration from \textit{PrefixSpan} and 
we can see that they indeed behave similarly. Although, for the dense iPRG dataset \textit{PrefixSpan} performs better than \textit{global-p.f.} and inversely for the large and sparse FIFA dataset. This might be due to implementation choices in the CP solver and \textit{PrefixSpan} software.



\paragraph{Analysis of the pattern quality}
Finally, we use our constraint-based framework to perform exploratory analysis of the Unix user datasets.
Table~\ref{tab:qual} shows different settings we tried and patterns we found interesting. Few constraints lead to too many patterns while more constrained settings lead to fewer and more interesting patterns.


\begin{table}[b]
  \centering
  \footnotesize
  \begin{tabular}{|l|c|c|c|}
    \hline
    setting & \# of patterns & interesting pattern      & comment                                   \\\hline
    ${\bf F_1}$   & 627                & -                        & Too many patterns                         \\\hline
    ${\bf F_2}$   & 512                & -                        & Long sequences of \lit{cd} and \lit{ls}   \\\hline
    ${\bf F_3}$   & 36                 & \seq{{\tt latex,bibtex,latex}} & User2 is using Latex to write a paper                          \\\hline
    ${\bf D_1}$   & 7                  & \seq{{\tt emacs}} & User2 uses $Emacs$, his/her collaborators use $vi$      \\\hline
    ${\bf D_2}$   & 9                  & \seq{{\tt quota, rm, ls, quota}}  & User is out of disc quota                 \\\hline
  \end{tabular}
  \caption{Patterns with various settings (User~2): ${\bf F_1}$: $minfreq = 5\%$, ~ ${\bf F_2}$: $~{\bf F_1} \land \minsize=3$, ~ ${\bf F_3}$: ${\bf F_2} \land \maxgap=2 \land \maxspan=5$, ~ ${\bf D_1}$: $minfreq=5\% \land \discriminant=8$ (w.r.t. all other users), ${\bf D_2}$: $minfreq=0.4\% \land \discriminant=8 \land member(\lit{quota})$}
  \label{tab:qual}
\end{table}

\section{Related work}
\label{sec:related}

The idea of mining patterns in sequences dates from earlier work by Agrawal et al. \cite{agrawal1995mining} shortly after their well-known work on frequent itemset mining~\cite{agrawal1994fast}. The problem introduced in \cite{agrawal1995mining} 
consisted of finding frequent sequences of {\em itemsets}; that is: sequences of sets included in a database of sequences of sets. 
Mining sequences of individual symbols was introduced later by \cite{mannila1997discovery}; 
the two problems are closely related and one can 
adapt one to the other 
\cite{wang2004bide}. Sequence mining was driven by the application of market basket analysis for customer data spread over 
multiple days. Other applications include bio-medical ones where a large number of DNA and protein sequence datasets are available (e.g. \cite{ye2007efficient}), or natural language processing where sentences can be represented as sequences of words~(e.g. \cite{DBLP:conf/kdd/TattiV12}).

Several specialised algorithm have addressed the problem of constrained sequence mining. The cSpade algorithm \cite{zaki2000sequence} for example is an extension of the Spade sequence mining algorithm~\cite{zaki2001spade} that supports constraints of type 1, 2 and 3. PrefixSpan~\cite{han2001prefixspan} mentions regular expression constraints too. The LCMseq algorithm~\cite{OhtaniKUA09} also supports a range of constraints, but does not consider all embeddings during search.
Other sequence mining algorithms have often focussed on constraints of type 4, and on closed sequence mining in particular. CloSpan~\cite{yan2003clospan} and Bide~\cite{wang2004bide} are both extentions of PrefixSpan to mine \textit{closed} frequent sequences. We could do the same in our CP approach by adding constraints after each solution found, following~\cite{dominance_dp,kemmarictai}.

Different flavors of sequence mining have been studied in the context of a generic framework, and constraint programming in particular. They all study constraints of type 1, 2 and 4. In \cite{coquery2012sat} the setting of sequence patterns with explicit wildcards in a single sequence is studied: such a pattern has a linear number of embeddings. 
As only a single sequence is considered, frequency is defined as the number of embeddings in that sequence, leading to a similar encoding to itemsets. This is extended in~\cite{JSS-13-3} to sequences of itemsets (with explicit wildcards over a single sequence). \cite{kemmarictai} also studies patterns with explicit wildcards, but in a database of sequences. Finally, \cite{metivierconstraint} considers standard sequences in a database, just like this paper; they also support constraints of type 3. The main difference is in the use of a costly encoding of the inclusion relation using non-deterministic automata and the inherent inability to use projected frequency. 

\section{Conclusion and discussion}
\label{sec:conclusions}
We have investigated a generic framework for sequence mining, based on constraint programming. The difficulty, compared to itemsets and sequences with explicit wildcards, is that the number of embeddings can be huge, while knowing that one embedding exists is sufficient.

We proposed two models for the sequence mining problem: one in which the exists-embedding relation is captured in a global constraint. The benefit is that the complexity of dealing with the existential check is hidden in the constraint. The downside is that modifying the inclusion relation requires modifying the global constraint; it is hence not generic towards such constraints.
We were able to use the same \textit{projected frequency} technique as well-studied algorithms such as PrefixSpan~\cite{han2001prefixspan}, by altering the global exists-embedding constraint and using a specialised search strategy. Doing this does amount to implementing specific propagators and search strategies into a CP solver, making the problem formulation not applicable to other solvers out-of-the-box. On the other hand, it allows for significant efficiency gains.

The second model exposes the actual embedding through variables, allowing for more constraints and making it as generic as can be. 
However, it has extra overhead and requires a custom two-phased search strategy. 


Our observations are not just limited to sequence mining. Other pattern mining tasks such as tree or graph mining also have multiple (and many) embeddings, hence they will also face the same issues with a reified exists relation. Whether a general framework exists for all such pattern mining problems is an open question.

\subsection*{Acknowledgments}
The authors would like to thank Siegfried Nijssen, Anton Dries and R\'emi Coletta for discussions on the topic, and the reviewers for their valuable comments. This work was supported by the European Commission under project FP7-284715 ``Inductive Constraint Programming'' and a Postdoc grant by the Research Foundation -- Flanders.


\section*{Appendix}
\appendix

\section{Branching with projected frequency}\label{app:branch}
We want to branch only over the symbols that are still \textit{frequent} in the prefix-projected sequences. Taking the current partially assigned sequence into account, after projecting this prefix away from each transaction, some transactions will be empty and others will have only some subset of its original symbols left.

For each propagator  ${\var C_i} \leftrightarrow \exists e ~\mbox{s.t.}~ {\var S} \sqsubseteq_e T_i$ we maintain the (monotonically decreasing) set of symbols for that transaction in a variable ${\var X_i}$. The propagator in Algorithm~\ref{alg:global_emp} needs just a one line addition, that is, after line~\ref{a1:endwhile} we add the following:

\begin{algorithmic}
\State{propagate by removing from ${\var X_i}$ all symbols not in $\langle T_i[pos_e]..T_i[|T_i|] \rangle$ except $\epsilon$}
\end{algorithmic}
which removes all symbols from ${\var X_i}$ that do not appear after the current prefix.

The brancher than first computes the local frequency of each symbol across all ${\var X_i}$, and only branches on the frequent ones.
Let $\theta$ be the minimum frequency threshold, then the branching algorithm is the following:
\begin{algorithm}
\algtext*{EndIf}
  \footnotesize 
\caption{local-frequency-brancher$({\var S}, {\var X})$}
\begin{algorithmic}[1]
\State{$pos_S \gets $ position of first unassigned variable in ${\var S}$}
\For{$s$ in ${\var S}[pos_S]$}
\State{$count \gets 0$}
\ForAll{${\var X_i}$}
\If{$s \in D({\var X_i})$} \Comment{symbol in domain of ${\var X_i}$}
\State{$count \gets count + 1$}
\EndIf
\EndFor
\If{$count >= \theta$}
\State{add \textit{branch-choice} '${\var S} = s$'}
\EndIf
\EndFor
\State{branch over all \textit{branch-choice}s (if any)}
\end{algorithmic}
\end{algorithm}%

\section{Decomposition with explicit embedding variables, modeling details}\label{app:decomp}
The decomposition consists of two constraints: the \posmatch constraint and the \isemb constraint.

\subsubsection{\posmatch formulation, part 1}
The first constraint needed to enforce \\ \posmatch is formally defined as follows:
\begin{align}
    \forall i \in 1, \ldots, n, \forall j \in 1,\ldots, |T_i|: &\quad ({\var S_j} = T_i[{\var E_{ij}}]) \lor ({\var E_{ij}} = |T_i|+1)
\end{align}
Instead of modeling this with a reified \textit{element} constraint, we can decompose the element constraint over all values in ${\var E_{ij}}$ except the \textit{no-match} value $|T_i|+1$:
\begin{align}
  \forall i \in 1, \ldots, n, \forall j \in 1,\ldots, |T_i|, \forall x \in 1\ldots |T_i|:
      {\var E_{ij}} = x \rightarrow {\var S_j} = T_i[x] 
\end{align}
Observe that in the above formulation $T_i[x]$ is a constant, so the reified ${\var S_j} = v$ expressions can be shared for all unique values of $v \in \Sigma$.

Furthermore, using half-reified constraints we need only one auxiliary variable for both ${\var E_{ij}} = x \rightarrow {\var B}$ and ${\var B} \rightarrow {\var S_j} = s$, where the latter can be shared for all unique values of $s \in \Sigma$. This leads to $O(n\cdot k\cdot k)$ half-reified constraints of the former type and $O(k\cdot k)$ auxiliary variables and half-reified constraints of the latter type, with $k = max_i(|T_i|)$. 

\subsubsection{\posmatch formulation, part 2}
The second constraint needed to enforce \\ \posmatch is:
\begin{align}
\forall i \in 1, \ldots, n, \forall j \in 2,\ldots, |T_i|: &\quad ({\var E_{i(j-1)}} < {\var E_{ij}}) \lor ({\var E_{ij}} = |T_i|+1)
\end{align}

Formulating this in CP would not perform any propagation until $|T_i|+1$ is removed from the domain of ${\var E_{ij}}$. However, one can see that the lower-bound on ${\var E_{i(j-1)}}$, when not equal to $|T_i|+1$, can be propagated to the lower-bound of ${\var E_{ij}}$.

Consider the following example: let $S=[\{B,C,\epsilon\}, \{A,B,C,\epsilon\}, \{A,B,C,\epsilon\}]$ and $T_1=[A,B,C]$ then $k=3$ and $D(E_1)=\{2,3,4\}, D(E_2)=\{2,3,4\}, D(E_3)=\{2,3,4\}$. However, because $min(D(E_1)) = 2$ we know that $E_2 \neq 2$ and similar for $E_3$. This leads to $E_3=\{4\}$, from which the \isemb propagator can derive that there is no embedding of the pattern in $T_1$. This is a quite common situation.

This propagation can be obtained with the following decomposition over all elements of the domain (except $|T_i|+1$):
\begin{align}
\forall i \in 1\ldots n, \forall j \in 1 \ldots |T_i|-1, \forall x \in 1 \ldots |T_i|: ({\var E_{ij+1}} = x) \rightarrow ({\var E_{ij}} < x)
\end{align}
However, this would require in the order $O(n\cdot k^2)$ reified constraints and auxiliary variables.

Instead, we use a simple modification of the binary inequality propagator $X < Y$ that achieves the same required result. This propagator always propagates the lower-bound of $X$ to $Y$, and as soon as $|T_i|+1 \notin Y$ it propagates like a standard $X < Y$ propagator.

There are $O(n\cdot k)$ such constraints needed and no auxiliary variables.

\subsubsection{\isemb formulation}

The constraint is the following:
\begin{align}
\forall i \in 1, \ldots, n:\quad 
    {\var C_i} \leftrightarrow \forall j \in 1,\ldots, |T_i|: ~ ({\var S_j} \neq \epsilon) \rightarrow ({\var E_{ij}} \neq |T_i|+1)
\end{align}

Across all transactions, the reified ${\var S_j} \neq \epsilon$ expressions can be shared.
$O(n\cdot k)$ such constraints and auxiliary variables are needed in total. For each transaction, the forall requires $|T_i|$ times 2 auxiliary variables, one for reifying ${\var E_{ij}} \neq |T_i| + 1$ and one for reifying the implication. This leads to an additional $O(n\cdot k)$ auxiliary variables and constraints, plus $n$ reified conjunction constraints.

\section{Sub-search for the existence of a valid ${\var E_{i}}$}
\label{app:subsearch}
For each transaction $i$ independently, we can search for a valid assignment of the $E_i$ variables. As soon as a valid one is found, the sub search can stop and propagate the corresponding assignment to the ${\var C_i}$ and ${\var E_{i}}$ variables of the master problem. 

The following pseudo-code describes how we implemented this scheme as a brancher in a copying solver (implementation for a trailing solver is similar):
\begin{algorithm}
\algtext*{EndIf}
  \footnotesize 
\caption{sub-search-brancher$({\var C}, {\var E})$}
\begin{algorithmic}[1]
\State{\textit{substate} $\gets$ copy of the current search state}
\ForAll{${\var C_i}$}
\State{remove all other branchers (e.g. variable/value orderings)}
\State{add to \textit{substate} the variable/value ordering that tries ${\var C_i} = $\textit{true} before ${\var C_i} = $\textit{false}}
\State{add to \textit{substate} as next variable/value ordering to search over all ${\var E_{ij}}$ variables in lexicographic order, trying their smallest value first}
\State{solve \textit{substate}}
\If{\textit{substate} has a solution}
\State{save \textit{substate}'s assignment of ${\var C_i}$ and ${\var E_{i}}$}
\Else
\State{fail the master problem} \Comment{When no valid ${\var C_i}$,${\var E_{i}}$ can be found}
\EndIf
\EndFor
\State{merge all saved assignments and have this as the only resulting \textit{branch-choice} for the master problem}
\end{algorithmic}
\end{algorithm}

In the above algorithm, for each transaction $i$ we enter the loop and remove all branchers, meaning that there are currently no search choices for the subproblem. We then force the sub-search to only search over ${\var C_i}$ and ${\var E_{i}}$, such that an assignment for ${\var C_i} = $\textit{true} is found first, if it exists. By removing all branchers at the start of loop, the next transaction's sub-search will not reconsider branching choices made in the previous sub-search.

As the master problem should not branch over any of the sub-search choices either, we merge all the assignments found by the sub-searches and present this as the only branch-choice for the master problem.

Using this sub-search-brancher, for each ${T_i}$ for which an embedding of ${\var S}$ in ${T_i}$ exists, ${\var C_i}$ will be \textit{true}. Only if no such embedding exists will ${\var C_i}$ be \textit{false}. This is the required behaviour for our constraint formulation.

\section{Projected frequency for explicit embedding variables}\label{app:projfreq}
We introduced the following constraint specification:
\begin{align}
      \forall j \in 1\ldots n, x \in \Sigma, \quad
      {\var S_j} = x \rightarrow |\{ i : {\var C_i} \wedge T_i[{\var E_{ij}}] = x \}| \ge \theta
\end{align}

A naive formulation of this expression would require reifying an element constraint ${\var B} \leftrightarrow T_i[{\var E_{ij}}] = x$. Instead, we will create element constraints $T_i[{\var E_{ij}}] = {\var A_{ij}}$, where ${\var A_{ij}}$ is an auxiliary integer variable. This leads to the following more efficient reformulation:
\begin{align}
    \label{eq:freq-char-reif2}
      \forall i \in 1\ldots n, j \in 1\ldots |T_i|, \quad &T_i[{\var E_{ij}}] = {\var A_{ij}} \\
      \forall i \in 1\ldots n, j \in 1\ldots |T_i|, x \in \Sigma, \quad &{\var S_j} = x \rightarrow |\{ i : {\var C_i} \wedge {\var A_{ij}} = x \}| \ge \theta
\end{align}
\optional{
This requires $O(n\cdot k)$ element constraints and auxiliary integer variables, $O(k\cdot |\Sigma|)$ half-reified linear inequality constraints that use an auxiliary variable, and $O(n\cdot |\Sigma|\cdot k^2)$ reified constraints and variables, for the ${\var T_i} \wedge {\var C_{ij}} = x$ expressions.
}

\bibliography{biblio}
\bibliographystyle{splncs03}
\end{document}